\documentclass{article}
\usepackage[preprint]{neurips_2025}

\usepackage[utf8]{inputenc} 
\usepackage[T1]{fontenc}    
\usepackage{hyperref}       
\usepackage{url}            
\usepackage{booktabs}       
\usepackage{amsfonts}       
\usepackage{nicefrac}       
\usepackage{microtype}      
\usepackage{xcolor}         
\usepackage{amsmath}
\usepackage{amsthm}
\usepackage{graphicx}
\usepackage[table]{xcolor}
\usepackage{amssymb}
\usepackage{multirow}
\hypersetup{
    colorlinks=true,
    linkcolor=red,
    citecolor=teal,
    urlcolor=cyan,
}
\newtheorem{proposition}{Proposition}
\newtheorem{definition}{Definition}
\usepackage{wrapfig}
\title{Visual-Guided Key-Token Regularization for Multimodal Large Language Model Unlearning}

\author{%
  Chengyi Cai$^1$, Zesheng Ye$^1$, Peike Li$^2$, Bo Han$^3$, Jianzhong Qi$^1$, Feng Liu$^1$ \\
  \quad$^1$The University of Melbourne
  \quad$^2$Google Research
  \quad$^3$Hong Kong Baptist University \\
  \texttt{fengliu.ml@gmail.com} 
}

\begin{document}

\maketitle

\begin{abstract}
Unlearning in Multimodal Large Language Models (MLLMs) prevents the model from revealing private information when queried about target images. Existing MLLM unlearning methods largely adopt approaches developed for LLMs. They treat all answer tokens uniformly, disregarding their varying importance in the unlearning process. Moreover, these methods focus exclusively on the language modality, disregarding visual cues that indicate key tokens in answers. In this paper, after formulating the problem of unlearning in multimodal question answering for MLLMs, we propose \textit{\textbf{Vi}sual-Guided \textbf{Ke}y-Token \textbf{R}egularization} (ViKeR). We leverage irrelevant visual inputs to predict ideal post-unlearning token-level distributions and use these distributions to regularize the unlearning process, thereby prioritizing key tokens. Further, we define key tokens in unlearning via \textit{information entropy} and discuss ViKeR’s effectiveness through token-level \textit{gradient reweighting}, which amplifies updates on key tokens. Experiments on MLLMU and CLEAR benchmarks demonstrate that our method effectively performs unlearning while mitigating forgetting and maintaining response coherence.

\end{abstract}

\section{Introduction}
\begin{figure}[t]
  \centering
  \includegraphics[width=\textwidth]{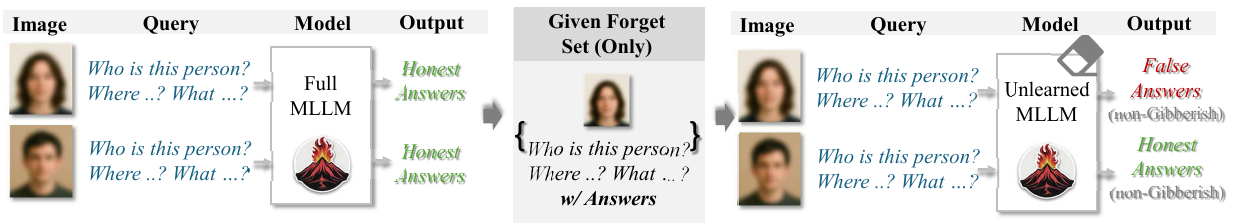}
    \caption{Problem formulation for unlearning. Given the forget set (i.e., target visual--question--answer triples to be unlearned), the MLLM is expected to forget the targeted content while preserving other knowledge and generating coherent responses.}
    \label{fig:problem_setting}
\end{figure}

Multimodal Large Language Models (MLLMs) have shown remarkable capabilities in vision-language understanding and generation by merging visual encoders and Large Language Model (LLM) backbones~\citep{mllm1,mllm2,mllm3}. Nevertheless, their tendency to memorize training data brings increasing concerns about privacy leakage and copyright violations~\citep{copyright1,copyright2}. To address these concerns, machine unlearning~\citep{unlearning} in MLLMs seeks to prevent the model from disclosing private information when queried about target persons~\citep{huo2025mmunlearner,MLLMU}, given the target set of visual--question--answer triples (referred to as the \textit{forget set}). Meanwhile, non-target information and the model’s capability to produce coherent answers should be preserved, as shown in Figure~\ref{fig:problem_setting}.

\begin{figure}[t]
    \centering
    \includegraphics[width=\textwidth]{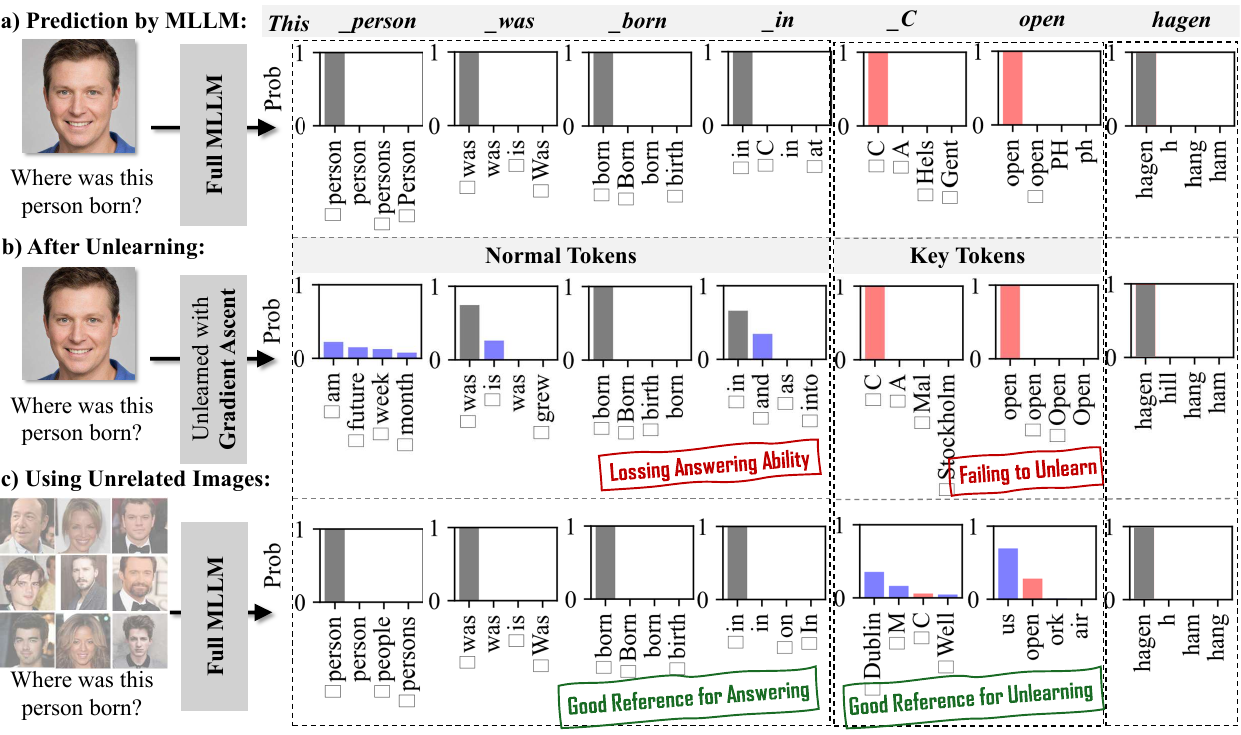}
    \caption{Probability distributions of the predicted answer tokens for a visual--question pair to be unlearned. a) Output of the full MLLM (i.e., LLM before unlearning). Gray indicates normal tokens, while red indicates key tokens. b) Output of MLLM after unlearning with GA, where normal tokens are forgotten instead of key tokens. c) Averaged output of the full MLLM, using irrelevant reference images as inputs, which approaches the ideal distribution for unlearning, and can be considered as a good reference for the unlearning process.}
    \label{fig:motivation}
\end{figure}

Research on MLLM unlearning is still at an early stage, and many existing approaches~\citep{MLLMU} are adapted from well-established LLM unlearning methods~\citep{ga,npo}. These methods typically assign a well-designed unlearning loss to the ground-truth answers corresponding to visual--question pairs in the forget set, discouraging the model from producing them. Such losses treat all tokens in the answer equally, which may instead cause less critical tokens that form the basic structure of a response without sensitive information (i.e., normal tokens) to be forgotten. As a result, the MLLM may lose its ability to generate coherent responses. Considering \textit{gradient ascent} (GA)~\citep{ga} as an example, the probability distributions of the predicted answer tokens for a visual--question pair (from the forget set of MLLMU~\citep{MLLMU}) before and after unlearning are shown in Figure~\ref{fig:motivation}(a) and Figure~\ref{fig:motivation}(b). Gray tokens indicate normal tokens, while red tokens indicate key tokens carrying critical information. It is observed that normal tokens are forgotten, leading to unreasonable token predictions, such as `This am'. However, key tokens such as `\text{\_}C' and `open' that form `Copenhagen' and reveal private information tend to be kept. This highlights the drawback of ignoring token-level importance.
However, automatically identifying key tokens in a sentence for unlearning remains a challenging problem. We observe that visual inputs can be effective in guiding key-token identification, an aspect that has been largely neglected by previous MLLM unlearning methods that focus solely on the language modality. We randomly choose 9 images from Celeb~\citep{celeb}, which are irrelevant to the forget set, to replace the image in Figure~\ref{fig:motivation}(a). These images, along with the same question, are fed into the full MLLM (i.e., the model before unlearning with full information), and the averaged token-level distributions are shown in Figure~\ref{fig:motivation}(c). It is observed that the predictions for normal tokens resemble those in Figure~\ref{fig:motivation}(a), while key tokens no longer convey private information, reflecting the ideal token probabilities after unlearning. Therefore, these distributions can serve as a reference to regularize the unlearning process.

Section~\ref{sec:problem_setting} formalizes the problem setting of MLLM unlearning, focusing on forgetting visual--question--answer triples. We further clarify that the objective is to remove the forget set while preserving non-target knowledge and MLLM’s ability to generate coherent responses, as shown in Figure~\ref{fig:problem_setting}. 
\clearpage
In Section~\ref{sec:method}, inspired by the observation in Figure~\ref{fig:motivation}, we propose a \textit{\textbf{Vi}sual-Guided \textbf{Ke}y-Token Regularization} (ViKeR) method for MLLM unlearning. ViKeR consists of two steps: visual-guided token distribution estimation and token-level regularization. Firstly, ideal token distributions are estimated using irrelevant visual inputs. Secondly, the estimated distributions are used to regularize the MLLM unlearning loss, which enables unlearning while preserving retention and response coherence.

In Section~\ref{sec:theory}, we use \textit{information entropy} to formally define key tokens. We then provide a discussion of ViKeR’s effectiveness by exploring its token-level \textit{gradient reweighting}, in comparison with GA.

In Section~\ref{sec:results}, experimental results on the MLLMU~\citep{MLLMU} and CLEAR~\citep{clear} benchmarks with LLaVA-7B~\citep{llava} demonstrate the effectiveness of ViKeR for MLLM unlearning. While achieving comparable unlearning performance, ViKeR outperforms existing unlearning methods in preserving non-target knowledge and maintaining output readability. The ablation studies, hyperparameter analyses, investigations of regularizers and reference images comprehensively investigate its performance and validate the soundness of the design.

In conclusion, both theoretical analysis and experimental results demonstrate the effectiveness of ViKeR. It offers new perspectives on improving MLLM unlearning through the use of visual modality and token-level importance.

\section{Related Works}
\textbf{LLM Unlearning.}
With the development of LLMs and the growing need for privacy protection~\citep{fantowards,yao2024survey,jang2023knowledge}, LLM unlearning has gradually emerged as a mature research direction. Recent studies cover a wide range of aspects, including LLM unlearning benchmarks~\citep{maini2024tofu,shimuse,li2024wmdp}, evaluation of methods~\citep{wangrethinking,wangtowards}, unified frameworks~\citep{dorna2025openunlearning}, and methodological improvements~\citep{jia2024soul,pawelczykcontext,kadhe2024split,wang2025gru,yang2025exploring}. 

One of the most well-known approaches is the \textit{gradient ascent} (GA)~\citep{ga}, which uses the negative log-likelihood typically employed in standard LLM training, but with the sign inverted as the loss function, allowing the model to forget the target data. In addition, \textit{direct preference optimization} (DPO)~\citep{rafailov2023direct} has also been applied to unlearning by treating responses such as “I don’t know” as preferred answers, encouraging the model to refuse to answer queries about private data. \textit{Negative preference optimization} (NPO)~\citep{npo}, adapts DPO by removing the preferred-answer component, satisfying cases when only non-preferred responses are available.

\textbf{MLLM Unlearning.}
Machine unlearning~\citep{bourtoule2021machine,cao2015towards,unlearning} has gradually been applied to MLLMs to remove harmful information~\citep{chen2025safeeraser} or private data~\citep{clear,MLLMU}. Currently, MLLM unlearning lacks a unified problem formulation: some studies focus on multi-person scenarios~\citep{chen2025auvic}, some on single images~\citep{li2024single}, some on the removal of harmful content~\citep{chen2025safeeraser}, and others on the elimination of visual modality-related answers while retaining language information~\citep{huo2025mmunlearner}. More discussion and evaluation involve hallucination mitigation~\citep{xing2024efuf}, safety alignment~\citep{chakraborty2024cross}, and attacks~\citep{zhang2025does}.

In this work, we focus on unlearning in multimodal question answering. To better align with practical applications, we restrict access during the unlearning stage to only the forget set, which is significantly smaller than the full dataset. Current MLLM unlearning methods sometimes involve internal architectural modifications, such as neuron pruning~\citep{liu2025modality}, neuron path editing~\cite{li2025cross}, or selective fine-tuning of specific weights~\citep{MLLMU}. However, considering architecture-agnostic unlearning loss design under our problem formulation, most available methods largely follow approaches from LLM unlearning~\citep{MLLMU}, often neglecting the varying importance of tokens and potential visual guidance.

\section{Preliminary}
\label{sec:problem_setting}
\textbf{Token-Level Probabilities and MLLM Training Loss.} We consider an MLLM parameterized by $\theta$, where the output of the image encoder is mapped by a projector into visual tokens and concatenated with LLM’s text tokens~\citep{llava}.
Let a visual--question--answer triple $s$ consist of an image $I$, a question $x$, and a ground-truth answer $y$, i.e., $s=(I, x, y)$. The answer $y$ is represented as a sequence of tokens $y=[y_1, y_2, \ldots, y_{|y|}]$. Let $y_{<i}$ denote the prefix $[y_1, \dots, y_{i-1}]$ of the answer $y$, the probability that the predicted $i$-th token $\hat y_i$ takes a specific value $v$ is defined as
\begin{equation}\label{eq:token_prob}
    p_{\theta}(v|I, x, i)\triangleq p(\hat y_i=v| I, x, y_{<i};\theta), 
\end{equation}
which is the output of the auto-regressive MLLM on each step for predicting the next token. Then the probability of answer $y$ is $p(y|I, x;\theta)=\prod_{i=1}^{|y|}p_{\theta}(y_i|I, x, i)$, i.e., the product of probabilities that the $i$-th predicted token $\hat y_i$ matches the ground truth token $y_i$. For a visual--question--answering task with training set $\mathcal D_{\rm full}=\{s^1,s^2, ..., s^{|\mathcal D|}\}$, the MLLM can be fine-tuned to obtain the full model $\theta_{\rm full}$ with loss
\begin{equation}\label{eq:loss}
   \mathcal L_{\rm NLL}(\mathcal D_{\rm full};\theta)=-\frac{1}{|\mathcal D|}\sum\nolimits_{s \in \mathcal D_{\rm full}}\log p(y|I, x;\theta),
\end{equation}
where $L_{\rm NLL}(\cdot; \cdot)$ denotes the negative log-likelihood.

\begin{figure*}[t]
    \centering
    \includegraphics[width=\linewidth]{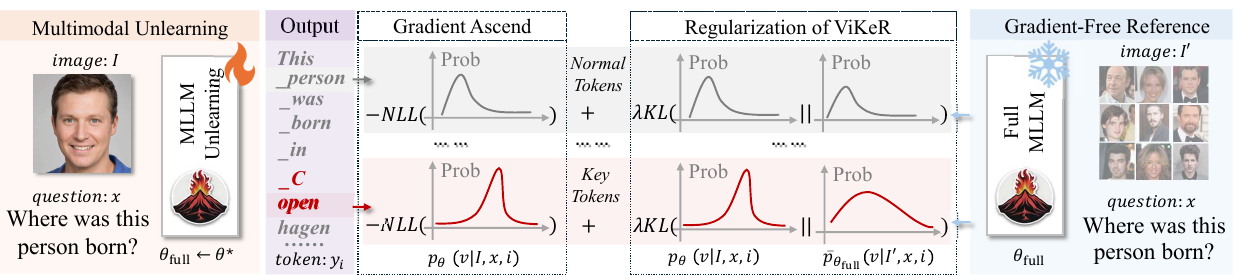}
    \caption{ViKeR Pipeline.
The left half shows GA, while the right depicts modifications of ViKeR. ViKeR has two stages: (1) \textit{visual-guided token distribution estimation} (in blue), where irrelevant visual inputs are fed into the pre-unlearning gradient-free full model to estimate ideal token distributions, and (2) \textit{token-level regularization} (in orange), using the predicted distribution to regularize GA.}
\label{fig:pipeline}
\end{figure*}
\textbf{Problem Formulation for MLLM Unlearning.} Similar to LLM unlearning, we define the forget set as $\mathcal D_{\rm f}=\{s^1_{\rm f}, s^2_{\rm f}, ..., s^{|\mathcal D_{\rm f}|}_{\rm f}\}\subset \mathcal{D}_{\rm full}$, consisting of visual--question--answer triples to be unlearned from $\theta_{\rm full}$. The size of these datasets typically satisfies $|\mathcal D_{\rm f}|\ll|\mathcal D_{\rm full}|$. During the unlearning phase, only $\theta_{\rm full}$ and $\mathcal D_{\rm f}$ are available, which reflects real-world usage scenarios. Then the goal of unlearning is to derive a model $\theta^*$ that simultaneously satisfies:
\begin{itemize}
    \item (\textit{Forgetting}). For any $(I, x, y) \in \mathcal D_{\rm f}$, $\theta^*$ no longer outputs the correct answer $y$ when queried with the visual--question pair $(I, x)$.

    \item (\textit{Retention}). For any $(I, x, y) \in \mathcal D_{\rm full} \setminus \mathcal D_{\rm f}$, the output of $\theta^*$ remains unchanged.

    \item (\textit{Coherence}). For any $(I, x, y) \in \mathcal D_{\rm full}$, the model $\theta^*$ preserves its ability to generate coherent, non-gibberish outputs for inputs $(I, x)$.
\end{itemize}

\textbf{Gradient Ascent (GA).} An intuitive approach is to reuse the negative log-likelihood in Eq.~\eqref{eq:loss} on $\mathcal D_{\rm f}$, but with the sign flipped, and use it as the unlearning loss: 
\begin{equation}\label{eq:ga}
    \mathcal L_{\rm GA}(\mathcal D_{\rm f};\theta)=-\mathcal L_{\rm NLL}(\mathcal D_{\rm f};\theta).
\end{equation}
Substituting Eq.~\eqref{eq:token_prob} and Eq.~\eqref{eq:loss} into Eq.~\eqref{eq:ga}, we can express the gradient of GA loss in terms of individual tokens:
\begin{equation}
    \nabla_{\theta} \mathcal L_{\rm GA}(\mathcal D_{\rm f};\theta)=\frac{1}{|\mathcal D_{\rm f}|}\sum\nolimits_{(I,x,y) \in \mathcal D_{\rm f}} \sum\nolimits_{i=1}^{|y|} \nabla_{\theta} \mathcal L_{\rm GA}(y_i;\theta, y, i), \notag
\end{equation}
where $\nabla_{\theta} \mathcal L_{\rm GA}(v;\theta, y, i) \triangleq \nabla_{\theta}\log p_{\theta}(v|I, x, i)$ is defined as the token-level gradient of the $i$-th token in answer $y$ being $v$.
See detailed derivation in Appendix~\ref{app:ga}. 

\textbf{Limitations of GA in MLLM unlearning.}
First, GA simply averages gradients over $\mathcal D_{\rm f}$ and assigns equal weight to all answer tokens, which leads to over-forgetting normal tokens and disrupted coherence, as illustrated in Figure~\ref{fig:motivation}(b).
Second, the GA loss in Eq.~\eqref{eq:ga} has the same form for MLLMs and text-only LLMs, treating the multimodal input $(I, x)$ as a flat token sequence and ignoring visual structure. Consequently, GA fails to leverage visual cues to distinguish identity-dependent tokens (i.e., key tokens) from those that should remain stable (i.e., normal tokens).

\section{Visual-Guided Key-Token Regularization}
\label{sec:method}
\textbf{Method Overview.} 
The pipeline of ViKeR, illustrated in Figure~\ref{fig:pipeline}, consists of two components: \textit{visual-guided token distribution estimation} and \textit{token-level regularization}.
The first part leverages irrelevant visual inputs (i.e., images of individuals not included in $\mathcal D_{\rm f}$) fed into the pre-unlearning full model to predict the ideal post-unlearning token distribution (as observed in Figure~\ref{fig:motivation}(c)) in a gradient-free manner.
The second component uses the predicted ideal distribution to regularize the GA unlearning process.

\textbf{Visual-Guided Token Distribution Estimation.} 
Shown in Figure~\ref{fig:motivation}(c), a set of unlabeled images of irrelevant individuals can be used to estimate the ideal post-unlearning token distribution for each generated answer token.

Assuming that $k$ images of irrelevant individuals from a reference set $\mathcal I'=\{I'_1, I'_2, ..., I'_k\}$ are used, the ideal distribution of the $i$-th token in an answer $y$ is estimated as:
\begin{equation}
    \hat{\mathcal R}^{y}_i (v) = \bar p_{\theta_{\rm full}}(v|\mathcal I', x,i)\triangleq\frac{1}{k} \sum\nolimits_{j=1}^k p_{\theta_{\rm full}}(v|I_j', x,i), \label{eq:ref_dis} 
\end{equation}
where $\hat{\mathcal R}^{y}_i$ is the distribution over tokens $v \in \mathcal V$, with $\mathcal V$ being the complete token set, and $p_{\theta_{\rm full}}(v|\cdot)$ is the probability of token $v$ as defined in Eq.~\eqref{eq:token_prob}, given the full model $\theta_{\rm full}$, a visual--question--answer triple $(I'_j, x, y)$ and index $i$.

It can be noticed that computing $\hat{\mathcal R}^{y}_i$ only requires forward passes through $\theta_{\rm full}$, being a gradient-free inference process. Therefore, it incurs minimal computational overhead. The hyperparameter $k$ is discussed in Section~\ref{sec:results}.

\textbf{Token-Level Regularization.}
Given an MLLM model $\theta$, a visual--question--answer triple $(I, x, y)\in D_{\rm f}$ and the token position $i$, the predicted distribution of the $i$-th token in $y$ is
\begin{equation}
    \hat{\mathcal Q}^{y}_i(v)  =  p_{\theta}(v|I, x,i), \label{eq:cur_dis} \notag
\end{equation}
as defined in Eq.~\eqref{eq:token_prob}. $\hat{\mathcal Q}^{y}_i$ is the distribution over $v \in \mathcal V$.

After ideal unlearning, we would like $\hat{\mathcal Q}^{y}_i$ to approximate $\hat{\mathcal R}^{y}_i$ for all $(I, x, y) \in \mathcal D_{\rm f}$ and positions $i$.
The intuition is that unlearning is achieved when, for any $(I, x, y) \in \mathcal{D}_{\rm f}$ fed into the MLLM, the resulting token
distributions resemble the expected distributions obtained by inputting the same question $x$ and identity-agnostic individuals.
It is supported by the observation in
Figure~\ref{fig:motivation}(c): across reference images from
$\mathcal I^\prime$, the averaged distributions
$\hat{\mathcal R}^{y}_i$ do not exhibit individual-specific peaks for
{\em key tokens} associated with private attributes, while remaining
sharp for {\em normal tokens}.

Therefore, the discrepancy between $\hat{\mathcal Q}^{y}_i$ and $\hat{\mathcal R}^{y}_i$ can be used to regularize the GA unlearning process. ViKeR employs the widely-used Kullback--Leibler (KL) divergence~\citep{kl} for regularization, so the unlearning loss can be formulated as:
\begin{equation}\notag
     \mathcal L_{\rm ViKeR}(\mathcal D_{\rm f};\theta)=-\mathcal L_{\rm NLL}(\mathcal D_{\rm f};\theta)+\frac{\lambda}{|\mathcal D_{\rm f}|} \cdot \sum\nolimits_{(I,x, y) \in \mathcal D_{\rm f}}\sum\nolimits_{i=1}^{|y|}\text{KL}(\hat{\mathcal R}^{y}_i||\hat{\mathcal Q^{y}_i}), \label{eq:viker_loss}
\end{equation}
where $\text{KL}(\cdot||\cdot)$ is the KL divergence, with a balance hyperparameter $\lambda\in [0,1]$, which is further discussed in Section~\ref{sec:results}. 

\section{Understanding ViKeR Through Token-Level Gradient Reweighting}
\label{sec:theory}
Let ${\mathcal R}^{y}_i$ denote the ideal post-unlearning token distribution at position $i$ in answer $y$, which is approximated in practice by  $\hat{\mathcal R}^{y}_i$ defined in Eq.~\eqref{eq:ref_dis}.
Then normal and key tokens in the ground truth answer $y$ are defined as follows. 

\begin{definition} \label{def:normal_token}
    (Normal Tokens.) \textit{Let $f_{\rm nrl}(\cdot)$ be an indicator function that marks whether a token is a normal token.
    We define a token $y_i$ as a normal token if its estimated ideal distribution $\mathcal{R}_i^y$ is sufficiently close to the Dirac distribution $\delta_{y_i}$ (i.e., a one-hot encoding at $y_i$). Formally, this is captured by the indicator function:$$f_{{\rm nrl}}(y_i) \triangleq \mathbb{I}\{ \mathcal{R}_i^y(y_i) \ge \tau \}$$where $\tau \in (0, 1]$ is a confidence threshold satisfying $\tau \rightarrow 1$.}
\end{definition}
For example, `\text{\_}person', `\text{\_}was', `\text{\_}born', `\text{\_}in', `hagen' in Figure~\ref{fig:motivation} can be normal tokens.
\begin{proposition}\label{prop:5.2}
    Let $\text{H}(\cdot)$ denote the information entropy~\citep{shannon1948mathematical}. Then for a normal token $y_i$: $$f_{\rm nrl}(y_i)=1 \Rightarrow \text{H}(\mathcal R^y_i) \rightarrow 0,$$proved in Appendix~\ref{app:prop5.2}. The ideal post-unlearning distribution of a normal token has (approximately) zero entropy.
\end{proposition}
Accordingly, key tokens can be defined distinctly:
\begin{definition}
(Key Tokens.) \textit{Let $f_{\rm key}(\cdot)$ be an indicator function that marks whether a token is a key token. Then $$f_{\rm key}(y_i) \triangleq \mathbb{I}\{\text{H}(\mathcal R^y_i)\geq \epsilon\},$$where $\epsilon$ denotes a user-specified threshold and is independent of the subsequent propositions.}
\end{definition}
For example, `\text{\_}C', `open' in Figure~\ref{fig:motivation} can be key tokens.

By definition, key tokens and normal tokens should be disjoint and do not cover all tokens, and they are treated with different importance during unlearning. Next, we differentiate the ViKeR loss in Eq.~\eqref{eq:viker_loss} to analyze its token-level gradients and compare them to those of GA. 
\begin{proposition} \label{prop:5.4}
Recall the token-level GA gradient  $\nabla_{\theta} \mathcal L_{\rm GA}(v;\theta, y, i)$ defined in Section~\ref{sec:problem_setting}. If $y_i$ is a normal token, then its token-level gradient under ViKeR can be written as a reweighted version of the GA gradient: 
\begin{align}\notag
 \nabla_{\theta} \mathcal L_{\rm ViKeR}&(v;\theta, y, i) = \begin{cases}
        (1-\lambda)\cdot \nabla_{\theta} \mathcal L_{\rm GA}(v;\theta, y, i) & \text{ if } v=y_i \\ 0 \cdot \nabla_{\theta} \mathcal L_{\rm GA}(v;\theta, y, i) & \text{ if } v \neq y_i \notag
\end{cases}, 
\end{align}
where $v$ is the value of token variable in $i$-th location of answer $y$. Detailed proof is in Appendix~\ref{app:prop5.4_5.5}. 
\end{proposition}
Thus, for normal tokens, the unlearning gradients are scaled by a factor $1 - \lambda$ (or set to zero for $v \neq y_i$), which reduces the magnitude of updates on these tokens and helps preserve the model’s ability to generate them.

\begin{proposition}\label{prop:5.5} Using the definition in Eq.~\eqref{eq:ref_dis}, the token-level gradient under ViKeR can be written, for any token (with detailed proof in Appendix~\ref{app:prop5.4_5.5}), as:
\begin{align}\notag
    \nabla_{\theta} \mathcal L_{\rm ViKeR}&(v;\theta, y, i)= 
    (\mathbb I \{v=y_i\}- \lambda \cdot \bar p_{\theta_{\rm full}}(v|\mathcal I', x,i)) \cdot 
    \nabla_{\theta} \mathcal L_{\rm GA}(v;\theta, y, i),\notag
\end{align}
with Proposition~\ref{prop:5.4} following as a special case.
\end{proposition}
Key tokens are given more priority compared with normal ones, as the reweighting scale is larger than $1-\lambda$ (of normal tokens) when $v=y_i$. For other candidate tokens $v \neq y_i$, the gradients are reweighted in the opposite direction to unlearning, which leads $\theta$ to favor other candidate tokens, also contributing to forgetting the target $y_i$.

\section{Experiments}
\label{sec:results}
\textbf{Benchmarks and Baselines.}
Experiments are conducted on the MLLMU~\citep{MLLMU} and CLEAR~\citep{clear} benchmarks using LLaVA-7B~\citep{llava} with LoRA~\citep{hu2022lora} fine-tuning, following \citep{MLLMU}. 
Benchmark details and evaluation metrics are provided in Appendix~\ref{app:benchmark} and Appendix~\ref{app:metric}. 
Baselines include GA~\citep{ga}, NPO~\citep{npo}, and IdkPO (i.e., using `I don't know ...' as preferred samples in DPO~\citep{rafailov2023direct}). 
All methods share the same training hyperparameters for a fair comparison. 
Baseline descriptions and implementation details are given in Appendix~\ref{app:baseline} and Appendix~\ref{app:imple}.

\begin{table*}[t]
\caption{Results of different methods on MLLMU, with the best in \textbf{bold} and ours \colorbox{gray!30}{highlighted}. All reported values are percentages (\%).}
\centering

\begin{small}
\resizebox{\textwidth}{!}{
\begin{tabular}{c|cc|cc|ccc|ccc}
\toprule
\multicolumn{1}{l}{}          & \multicolumn{2}{|c|}{Forget}                              & \multicolumn{2}{c|}{Generalization}                      & \multicolumn{3}{c|}{Retain}                                                             & \multicolumn{3}{c}{Real}                                                              \\
\midrule
10\% Task        & Acc                          & Gib↑                    & Acc                          & Gib↑                    & Rouge↑                       & BLEU↑                        & Gib↑                    & Rouge↑                       & BLEU↑                       & Gib↑                    \\
\midrule
{\color[HTML]{A6A6A6} Origin} & {\color[HTML]{A6A6A6} 25.2} & {\color[HTML]{A6A6A6} -} & {\color[HTML]{A6A6A6} 30.8} & {\color[HTML]{A6A6A6} -} & {\color[HTML]{A6A6A6} -}    & {\color[HTML]{A6A6A6} -}    & {\color[HTML]{A6A6A6} -} & {\color[HTML]{A6A6A6} -}    & {\color[HTML]{A6A6A6} -}   & {\color[HTML]{A6A6A6} -} \\
{\color[HTML]{A6A6A6} Full}   & {\color[HTML]{A6A6A6} -}    & {\color[HTML]{A6A6A6} -} & {\color[HTML]{A6A6A6} -}    & {\color[HTML]{A6A6A6} -} & {\color[HTML]{A6A6A6} 49.7} & {\color[HTML]{A6A6A6} 23.1} & {\color[HTML]{A6A6A6} -} & {\color[HTML]{A6A6A6} 22.6} & {\color[HTML]{A6A6A6} 6.1} & {\color[HTML]{A6A6A6} -} \\
GA                            & \textbf{25.3}\scriptsize{±2.9}          & 83.7\scriptsize{±2.1}                      & \textbf{30.1}\scriptsize{±2.5}                & 78.5\scriptsize{±2.7}                      & 20.6\scriptsize{±2.9}                         & 3.0\scriptsize{±0.7}                          & 81.3\scriptsize{±3.5}                      & 17.5\scriptsize{±2.1}                         & 2.4\scriptsize{±0.6}                        & 87.8\scriptsize{±2.6}                     \\
NPO                           & 30.8\scriptsize{±1.7}                        & 89.9\scriptsize{±0.9}                     & 33.1\scriptsize{±1.4}                        & 80.9\scriptsize{±1.9}                     & 29.7\scriptsize{±1.1}                        & 5.7\scriptsize{±0.5}                         & 89.9\scriptsize{±0.8}                     & 23.2\scriptsize{±0.6}                        & 4.6\scriptsize{±0.2}                        & 91.2\scriptsize{±0.5}                     \\
IdkPO                           & 36.9\scriptsize{±1.3}                        & 94.4\scriptsize{±0.2}                     & 36.5\scriptsize{±0.2}                        & 73.2\scriptsize{±1.4}                     & 5.7\scriptsize{±0.5}                         & 1.3\scriptsize{±0.2}                         & 91.7\scriptsize{±0.2}                     & 4.2\scriptsize{±0.1}                         & 0.6\scriptsize{±0.0}                        & 86.3\scriptsize{±0.1}                     \\
\rowcolor{gray!30} ViKeR                        & 30.4\scriptsize{±3.6}                        & \textbf{95.4}\scriptsize{±0.3}            & \textbf{30.1}\scriptsize{±1.9}               & \textbf{88.2}\scriptsize{±3.3}            & \textbf{32.4}\scriptsize{±1.4}               & \textbf{6.9}\scriptsize{±0.5}                & \textbf{94.6}\scriptsize{±0.4}            & \textbf{24.6}\scriptsize{±1.1}               & \textbf{4.7}\scriptsize{±0.4}               & \textbf{93.7}\scriptsize{±0.2}            \\
\midrule
15\% Task        & Acc                          & Gib↑                    & Acc                          & Gib↑                    & Rouge↑                       & BLEU↑                        & Gib↑                    & Rouge↑                       & BLEU↑                       & Gib↑                    \\
\midrule
{\color[HTML]{A6A6A6} Origin} & {\color[HTML]{A6A6A6} 28.0} & {\color[HTML]{A6A6A6} -} & {\color[HTML]{A6A6A6} 29.6} & {\color[HTML]{A6A6A6} -} & {\color[HTML]{A6A6A6} -}    & {\color[HTML]{A6A6A6} -}    & {\color[HTML]{A6A6A6} -} & {\color[HTML]{A6A6A6} -}    & {\color[HTML]{A6A6A6} -}   & {\color[HTML]{A6A6A6} -} \\
{\color[HTML]{A6A6A6} Full}   & {\color[HTML]{A6A6A6} -}    & {\color[HTML]{A6A6A6} -} & {\color[HTML]{A6A6A6} -}    & {\color[HTML]{A6A6A6} -} & {\color[HTML]{A6A6A6} 50.0} & {\color[HTML]{A6A6A6} 24.2} & {\color[HTML]{A6A6A6} -} & {\color[HTML]{A6A6A6} 22.6} & {\color[HTML]{A6A6A6} 6.1} & {\color[HTML]{A6A6A6} -} \\
GA                            & \textbf{0.0}\scriptsize{±0.0}                         & 0.0\scriptsize{±0.0}                       & \textbf{0.0}\scriptsize{±0.0}                          & 0.0\scriptsize{±0.0}                       & 0.1\scriptsize{±0.0}                          & 0.0\scriptsize{±0.0}                          & 0.0\scriptsize{±0.0}                       & 0.1\scriptsize{±0.0}                          & 0.0\scriptsize{±0.0}                         & 0.0\scriptsize{±0.0}                       \\			
NPO                           & 7.1\scriptsize{±1.3}                          & 50.8\scriptsize{±6.8}                      & 12.0\scriptsize{±1.1}                         & 57.9\scriptsize{±2.0}                      & 11.3\scriptsize{±0.7}                         & 0.9\scriptsize{±0.1}                          & 51.4\scriptsize{±5.0}                      & 9.1\scriptsize{±0.6}                          & 0.7\scriptsize{±0.1}                         & 51.0\scriptsize{±3.8}                      \\
IdkPO                           & 34.7\scriptsize{±0.9}                         & 92.1\scriptsize{±0.6}                      & 37.2\scriptsize{±0.3}                         & 82.2\scriptsize{±1.9}                      & 5.2\scriptsize{±0.2}                          & 0.6\scriptsize{±0.0}                          & 93.6\scriptsize{±0.6}                      & 4.0\scriptsize{±0.2}                          & 0.4\scriptsize{±0.0}                         & 91.3\scriptsize{±1.3}                      \\
\rowcolor{gray!30}
ViKeR                        & 32.0\scriptsize{±3.1}                & \textbf{93.2}\scriptsize{±0.7}             & 32.4\scriptsize{±3.4}                & \textbf{83.1}\scriptsize{±0.8}             & \textbf{52.7}\scriptsize{±1.8}                & \textbf{22.0}\scriptsize{±1.7}                & \textbf{94.6}\scriptsize{±0.0}             & \textbf{33.2}\scriptsize{±0.6}                & \textbf{10.3}\scriptsize{±0.2}               & \textbf{92.0}\scriptsize{±0.7}             \\
\bottomrule
\end{tabular}}
\end{small}
\vspace{-15pt}
\label{tab:res_mllmu}
\end{table*}

\textbf{Comparison Results.} The results on MLLMU are shown in Table~\ref{tab:res_mllmu}. Methods are evaluated along `\textit{Forget}', `\textit{Generalization}', `\textit{Retain}', and `\textit{Real}'.
`\textit{Forget}' reports results on the forget set, while `\textit{Generalization}' evaluates the set from different views; both are measured using classification accuracy (ACC) for unlearning.
`\textit{Retain}' corresponds to synthetic persons to be retained, and `\textit{Real}' refers to real persons; both are evaluated using ROUGE~\citep{ROUGE} and BLEU~\citep{bleu} to measure information retention.
GIB~\citep{gibberish} is used to assess whether the generated responses are gibberish.
`Origin' and `Full' denote the MLLMs before and after training on $\mathcal D_{\rm full}$, respectively, and serve as references for unlearning and retaining performance.
It can be observed from Table~\ref{tab:res_mllmu} that GA and NPO show a clear degradation in GIB, while IdkPO substantially harms retention, resulting in low ROUGE and BLEU scores.
In contrast, ViKeR achieves forgetting accuracy close to the original MLLM while maintaining the best ROUGE, BLEU, and GIB performance. Its advantage becomes more pronounced in the task with more forgetting data (i.e., 15\%), yielding ROUGE gains of +41.4\% /+24.1\% and BLEU gains of +21.1\% / +9.6\% on `\textit{Retain}' and `\textit{Real}', respectively. 

\begin{wraptable}{l}{0.57\textwidth}
\caption{Results on the CLEAR dataset (\%, best is in \textbf{bold} and ours is \colorbox{gray!30}{highlighted}).}
\centering

\begin{small}
\begin{tabular}{>{\centering}m{0.8cm}|>{\centering}m{0.5cm}>{\centering}m{0.7cm}|>{\centering}m{0.5cm}>{\centering}m{0.7cm}|>{\centering}m{0.85cm}m{0.85cm}}
\toprule
                        & \multicolumn{2}{c|}{Forget} & \multicolumn{2}{c|}{Retain} & \multicolumn{2}{c}{Real QA} \\
\midrule
           10\% Task             & Rec↓         & Gib↑       & Rec↓        & Gib↑        & Acc (Aut)↑   & Acc (Wor)↑  \\
\midrule
GA     & \textbf{0.00 }        & 11.60       & 0.00        & 11.27        & 50.55  & 47.77 \\
                        & \scriptsize ±0.00        & \scriptsize ±9.92       & \scriptsize ±0.00       &\scriptsize ±10.27       & \scriptsize ±1.74  & \scriptsize ±0.13 \\
NPO    & 0.18         & \textbf{94.16}       & 0.80        & 93.31        & 69.98  & 48.41 \\
                        &\scriptsize ±0.12        &\scriptsize ±0.66       &\scriptsize ±0.17       &\scriptsize ±0.70        &\scriptsize ±0.62  &\scriptsize ±0.13 \\
IdkPO  & \textbf{0.00 }        & 93.33       & 0.65        & 93.35        & 75.28  & \textbf{49.95} \\
                        &\scriptsize ±0.00        &\scriptsize ±0.32       &\scriptsize ±0.05       &\scriptsize ±0.49        &\scriptsize ±0.31  &\scriptsize ±0.13 \\
\rowcolor{gray!30}          ViKeR & 0.62         & 92.98       & \textbf{4.21}        & \textbf{93.42}        & \textbf{76.60}  & 48.96 \\
\rowcolor{gray!30}               &\scriptsize ±0.12        & \scriptsize ±0.88       &\scriptsize ±1.49       &\scriptsize ±0.80        &\scriptsize ±0.31  &\scriptsize ±0.34 \\
\bottomrule
\end{tabular}
\end{small}
\label{tab:res_clear}
\end{wraptable}
Results on the CLEAR dataset are shown in Table~\ref{tab:res_clear}, evaluated from three perspectives. `\textit{Forget}' and `\textit{Retain}' measure unlearning and retaining via identity recognition accuracy (REC) with GIB used in gibberish-judging, while `\textit{Real QA}' assesses answer accuracy (ACC) on real authors (AUT) and world (WOR) images. It can be observed that ViKeR maintains answer accuracy on real-world questions and fluent language generation (i.e., a high GIB) on par with SOTA. Moreover, it substantially improves retention (+3.41\%) while minimally affecting unlearning (+0.48\%).
Samples of visual--question--answer triples from MLLMU and CLEAR are provided in Appendix~\ref{app:sampe_answers}, showing responses after unlearning by different methods.

\begin{figure*}[t]
    \centering
    \includegraphics[width=0.98\linewidth]{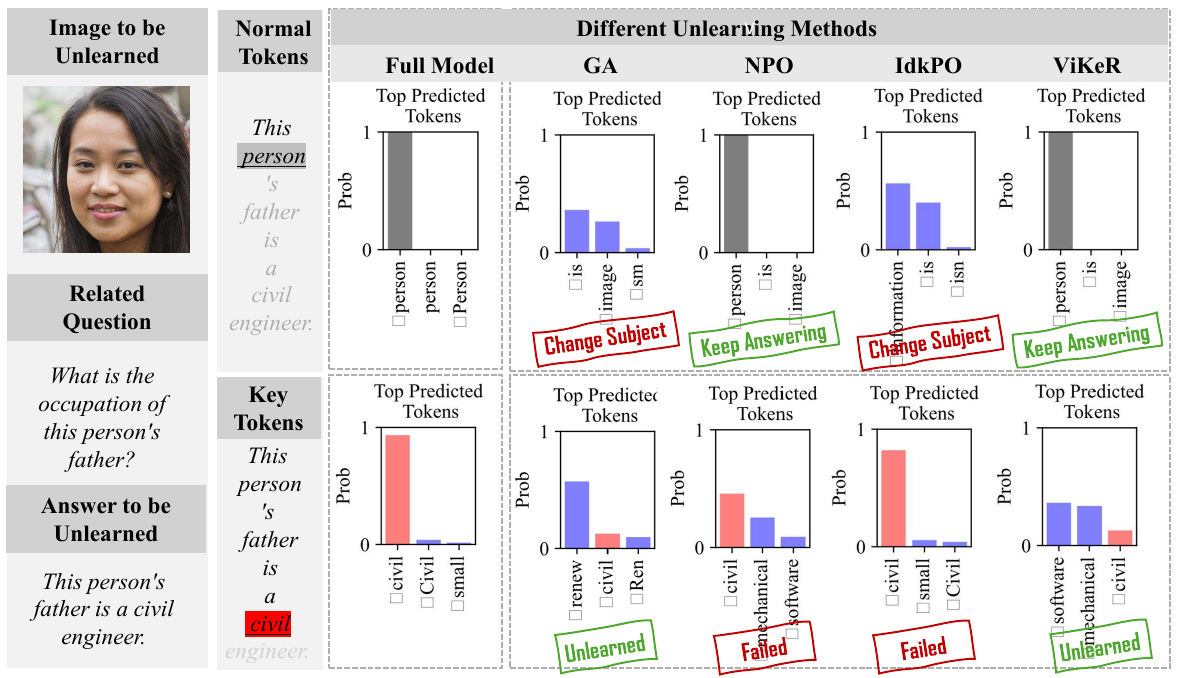}
    \caption{Visualization results of token distribution after unlearning with different methods. Compared to other methods, ViKeR preserves the prediction of normal tokens (such as `\text{\_}person') while successfully unlearning key tokens (such as `\text{\_}civil').}
    \vspace{-10pt}
    \label{fig:visual}
\end{figure*}

\textbf{Visualization Results of Token Distribution.} Figure~\ref{fig:visual} illustrates an example from the forget set, including the image, the associated question, and the answer. We visualize the predicted token distributions after unlearning with different methods, using the normal token `\text{\_}person' (gray bars) and the key token `\text{\_}civil' (red bars) as representative cases. The results show that GA significantly perturbs the distribution of normal tokens, leading to responses that drift away from the original question. While NPO maintains the ability to continue answering the question, it does not fully suppress the probability mass of the key token, and private information may still be leaked given the preceding context. IdkPO exhibits similar issues, including distortion of normal token predictions and failure in unlearning the key token. In contrast, our ViKeR effectively eliminates the key token while preserving coherent normal answers. More visualization results are illustrated in Appendix~\ref{app:visual}.

\textbf{Ablation Studies.} Table~\ref{tab:ablation} reports the ablation studies. `W/o Reg' denotes removing the second term in Eq~\eqref{eq:viker_loss}, i.e., the regularization term. `W/o GA' indicates removing the first term, i.e., the positive log-likelihood term. `W/o Vis' refers to removing the visual guidance and applying regularization solely based on the image’s initial token distribution, i.e., using $p_{\rm full}(v|I, x, i)$ as $\hat {\mathcal R}^y_i$ in Eq~\eqref{eq:ref_dis}.

It can be observed that `w/o Reg' degenerates to GA, leading to pronounced forgetting as well as a noticeable degradation in response coherence. This is evidenced by the lower ROUGE and BLEU scores on the `\textit{Retain}' and `\textit{Real}' sets, together with a reduced GIB score. In contrast, `w/o GA' weakens the unlearning capability, resulting in an unexpected increase in ACC on the `\textit{Forget}' and `\textit{Generalization}' sets. These results indicate that both terms in Eq~\eqref{eq:viker_loss} are indispensable, validating the design of the token-level regularization. Although `w/o Vis' relatively preserves response coherence, it still exhibits a significant drop in retaining performance, suggesting that our visual-guided token distribution estimation is likewise necessary. Please refer to Appendix~\ref{app:ablation} for more results.

\begin{table}[t]
\caption{Results of ablation studies on MLLMU, with the best in \textbf{bold} and ours \colorbox{gray!30}{highlighted}(\%).}
\vspace{-5pt}
\centering

\begin{small}
\resizebox{\textwidth}{!}{
\begin{tabular}{c|cc|cc|ccc|ccc}
\toprule
\multicolumn{1}{l}{}          & \multicolumn{2}{|c|}{Forget}                              & \multicolumn{2}{c|}{Generalization}                      & \multicolumn{3}{c|}{Retain}                                                             & \multicolumn{3}{c}{Real}                                                              \\
\midrule
10\% Task        & Acc                          & Gib↑                    & Acc                          & Gib↑                    & Rouge↑                       & BLEU↑                        & Gib↑                    & Rouge↑                       & BLEU↑                       & Gib↑                    \\
\midrule
{\color[HTML]{A6A6A6} Origin} & {\color[HTML]{A6A6A6} 25.2} & {\color[HTML]{A6A6A6} -} & {\color[HTML]{A6A6A6} 30.8} & {\color[HTML]{A6A6A6} -} & {\color[HTML]{A6A6A6} -}    & {\color[HTML]{A6A6A6} -}    & {\color[HTML]{A6A6A6} -} & {\color[HTML]{A6A6A6} -}    & {\color[HTML]{A6A6A6} -}   & {\color[HTML]{A6A6A6} -} \\
{\color[HTML]{A6A6A6} Full}   & {\color[HTML]{A6A6A6} -}    & {\color[HTML]{A6A6A6} -} & {\color[HTML]{A6A6A6} -}    & {\color[HTML]{A6A6A6} -} & {\color[HTML]{A6A6A6} 49.7} & {\color[HTML]{A6A6A6} 23.1} & {\color[HTML]{A6A6A6} -} & {\color[HTML]{A6A6A6} 22.6} & {\color[HTML]{A6A6A6} 6.1} & {\color[HTML]{A6A6A6} -} \\
w/o Reg                            & \textbf{25.3}\scriptsize{±2.9}          & 83.7\scriptsize{±2.1}                      & \textbf{30.1}\scriptsize{±2.5}                & 78.5\scriptsize{±2.7}                      & 20.6\scriptsize{±2.9}                         & 3.0\scriptsize{±0.7}                          & 81.3\scriptsize{±3.5}                      & 17.5\scriptsize{±2.1}                         & 2.4\scriptsize{±0.6}                        & 87.8\scriptsize{±2.6}                     \\
w/o GA                           & 44.9\scriptsize{±0.2}                        & 89.5\scriptsize{±1.0}                     & 39.5\scriptsize{±0.5}                        & 73.7\scriptsize{±1.5}                     &41.8 \scriptsize{±1.0}                        & 16.3 \scriptsize{±0.5}                         &87.3 \scriptsize{±0.8}                     & 20.1 \scriptsize{±0.4}                        & 5.8	\scriptsize{±0.1}                        & 74.2 \scriptsize{±1.0}                     \\
w/o Vis                          &26.7 \scriptsize{±3.8}                        & 90.1\scriptsize{±0.4}                     & 30.9\scriptsize{±2.6}                        & 85.3	\scriptsize{±0.3}                     & 28.2	\scriptsize{±1.9}                         & 5.0	\scriptsize{±0.6}                         & 90.1	\scriptsize{±0.5}                     & 21.5	\scriptsize{±1.6}                         & 3.7	\scriptsize{±0.7}                        &91.7 \scriptsize{±0.2}                     \\
\rowcolor{gray!30} ViKeR                        & 30.4\scriptsize{±3.6}                        & \textbf{95.4}\scriptsize{±0.3}            & \textbf{30.1}\scriptsize{±1.9}               & \textbf{88.2}\scriptsize{±3.3}            & \textbf{32.4}\scriptsize{±1.4}               & \textbf{6.9}\scriptsize{±0.5}                & \textbf{94.6}\scriptsize{±0.4}            & \textbf{24.6}\scriptsize{±1.1}               & \textbf{4.7}\scriptsize{±0.4}               & \textbf{93.7}\scriptsize{±0.2}            \\
\bottomrule
\end{tabular}}
\end{small}
\label{tab:ablation}
\vspace{-7pt}
\end{table}

\begin{figure}[htbp]
\centering
\begin{minipage}{0.51\textwidth}
  \centering
  \includegraphics[width=\linewidth]{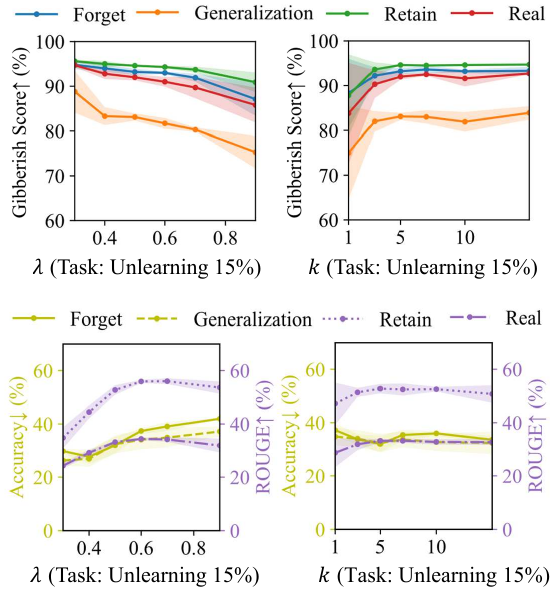}
    \caption{Hyperparameter analysis of $\lambda$ and $k$.}
    \label{fig:hyper}
\end{minipage}
\hfill
\begin{minipage}{0.47\textwidth}
  \centering
  \includegraphics[width=\linewidth]{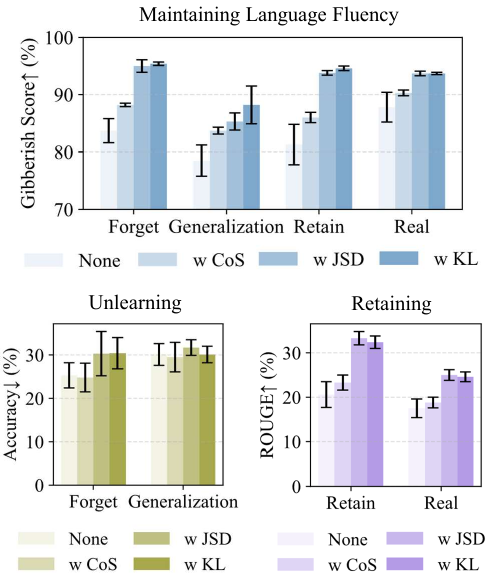}
    \caption{Different regularizers on 10\% Task.}
    \label{fig:aggre}
\end{minipage}
\vspace{-8pt}
\end{figure}

\textbf{Hyperparameter Analysis.} Figure~\ref{fig:hyper} shows the effects of the regularization strength $\lambda$ and the number of irrelevant visual inputs $k$ used to estimate $\hat {\mathcal R}^y_i$. The first row reports GIB, measuring the coherence of MLLM responses after unlearning, while the second row shows unlearning and retaining performance, evaluated by ACC and ROUGE, respectively.

As the regularization strength increases, retaining performance improves, but unlearning effectiveness degrades. When $\lambda$ is relatively large, the GIB score decreases and becomes unstable, with larger variance. We set $\lambda=0.5$ for the 15\% tasks as a balanced choice. In contrast, the number of irrelevant visual inputs $k$ has a smaller impact on the results. With fewer reference images, the performance is unstable and shows higher variance. When $k \geq 5$, the metrics become stable. We adopt $k=5$ in our experiments. Detailed results are included in Appendix~\ref{app:hyper}.

\textbf{Results of Different Regularization.} ViKeR employs KL divergence to regularize the distance between $\hat {\mathcal R}^y_i$ and $\hat {\mathcal Q}^y_i$, while other similarity measures can also be used as regularizers. Figure~\ref{fig:aggre} presents the results using cosine similarity (CoS)~\citep{lin2002divergence} and Jensen–Shannon divergence (JSD)~\citep{salton1975vector}. 

Compared to the setting without regularization, CoS substantially improves language coherence, but shows limited gains in unlearning and retaining performance. The results of JSD are nearly identical to those of KL divergence. However, from a computational perspective, JSD is approximately twice as expensive as KL divergence. Therefore, adopting KL divergence as the regularization term is a more reasonable choice. Detailed results are included in Appendix~\ref{app:aggregatio}.

\begin{figure*}
    \centering
    \includegraphics[width=\linewidth]{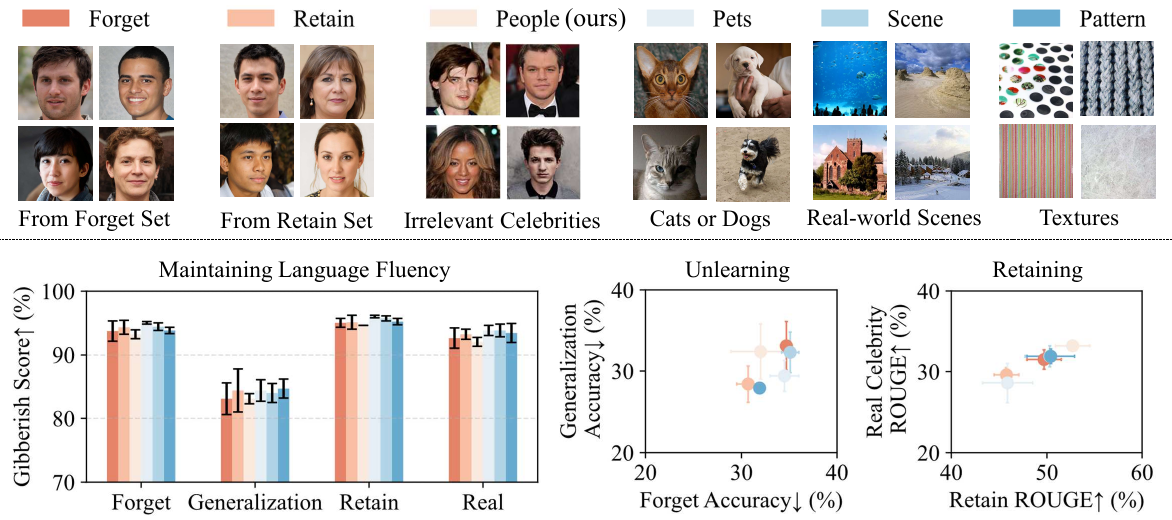}
    \caption{Results of ViKeR on MLLMU (15\% Task) when using different kinds of images as the reference set $\mathcal I'$, other than the celebrities.}
    \label{fig:ref}
    \vspace{-10pt}
\end{figure*}

\textbf{Results of Different Visual References.}
To study the impact of reference image set $\mathcal I'$, we further experiment with images from the forget set (`forget'), the retain set (`retain'), and other categories, including cats or dogs (`pets'), real-world scenes (`scene'), and different textures (`pattern'), aside from the irrelevant celebrities (`peoples') we use in ViKeR. The results are in Figure~\ref{fig:ref} and Appendix~\ref{app:reference}.

Types of references have little effect on GIB, indicating that language coherence is preserved regardless of $\mathcal I'$. However, unlearning and retaining performance vary. Using `pattern' and `retain' images yields the best unlearning performance, likely due to their relatively consistent internal distributions and large discrepancy from the forget set. In contrast, `forget' and `scene' images are the worst, as `forget' images contain privacy information to unlearn, and `scene' images exhibit high diversity, both of which hinder accurate ideal distribution estimation.
For retaining performance, `retain' and `pet' images lead to noticeable degradation, possibly because their estimated distributions contain task-related or human-like information that is also suppressed during unlearning. In contrast, using `people' images achieves a better balance between unlearning and retaining.

\section{Conclusion}
We propose ViKeR for MLLM unlearning with improved retention and response coherence. It leverages irrelevant visual inputs to estimate the ideal token distribution, which constrains the unlearning process via token-level regularization. Its efficacy is validated theoretically and empirically.

\bibliography{ViKeR}
\bibliographystyle{plain}
\clearpage
\appendix

\section{More Preliminary Knowledge}
\subsection{Benchmark Information}
\label{app:benchmark}
MLLMU~\citep{MLLMU} is a dataset consisting of 500 fictional characters and 153 public figures. Each character is associated with at least seven visual--question--answer triples and seven text-only question--answer pairs. The evaluation protocol includes four aspects: \textit{Forget}, \textit{Generalization}, \textit{Retain}, and \textit{Real}, corresponding to characters to be unlearned, alternative images of the same characters, characters to be retained, and real-person question--answer pairs, respectively. Since this work focuses on vision-language tasks, we only use visual--question--answer triples for our experiments.

CLEAR~\citep{clear} is a dataset consisting of 200 fictional characters and 3,700 images paired with question--answer annotations. The evaluation primarily considers four aspects: \textit{Forget}, \textit{Retain}, \textit{Questions about real authors}, and \textit{Questions about the real world}. For the \textit{Forget} and \textit{Retain} settings, we focus on person identification, where the question "What is the name of the person in the image" is used, along with ten different name-containing answers for each character designated for unlearning.

\subsection{Metrics Information}
\label{app:metric}
The metrics used in this work include the following.
\begin{itemize}
    \item \textbf{ACC} (\%), which measures accuracy in a multiple-choice setting, where the MLLM is prompted to select a single letter corresponding to one of the provided options. The final score is computed as the proportion of correct answers. Notably, whether a lower ACC necessarily indicates better unlearning performance remains debatable. A model may achieve zero accuracy by losing its ability to answer the question properly, for example, by producing irrelevant characters or refusing to respond, which does not represent desirable unlearning behavior. Therefore, we additionally report the accuracy of the MLLM before being trained on data containing the forget set as a reference point.
    \item \textbf{REC} (\%) is a metric specifically designed to evaluate whether a person in an image is correctly identified. When the model is asked who the person in the image is, the prediction is considered correct if the generated answer contains the true name of the person. A lower REC score indicates better unlearning performance or worse retaining performance, while a higher REC score reflects the opposite.
    \item \textbf{GIB} (\%) is a classifier~\citep{gibberish} designed to assess whether the responses generated by LLMs or MLLMs are gibberish. It categorizes outputs into four classes: \textit{Noise}, \textit{Word Salad}, \textit{Mild Gibberish}, and \textit{Clean}, and achieves an accuracy of 97.36\% on a multi-class text classification task. In this work, we use the predicted probability of the \textit{Clean} label and report the average across all responses as the final GIB score. A higher GIB score indicates cleaner and more coherent responses, while a lower score suggests outputs closer to gibberish.
    \item \textbf{ROUGE-L} (\%) is a metric that evaluates the similarity between a generated response and a reference answer based on the length of their longest common subsequence (LCS)~\citep{ROUGE}. It computes precision as the LCS length divided by the length of the generated response, and recall as the LCS length divided by the length of the reference answer. The final ROUGE-L score is obtained by combining precision and recall via an F-measure. We report the average ROUGE-L score across all responses. A higher ROUGE-L score indicates better content preservation and higher similarity to the reference answer.
    \item \textbf{BLEU} (\%) is a precision-based metric that evaluates the overlap between a generated response and a reference answer at the n-gram level~\citep{bleu}. For each n-gram size, it computes the number of n-grams in the generated response that also appear in the reference, divided by the total number of n-grams in the generated response (clipped precision). The final BLEU score is calculated as the geometric mean of the clipped precisions across different n-gram sizes, multiplied by a brevity penalty to penalize overly short outputs. We report the average BLEU score across all responses. A higher BLEU score indicates more accurate and fluent generation, while a lower score reflects poorer correspondence to the reference answer.

\end{itemize}

\subsection{Baseline Information}
\label{app:baseline}
\textbf{GA.} As has been discussed in detail in this paper, the loss of GA can be formulated as $$\mathcal{L}_{\rm GA}(D_{\rm f};\theta)=-\mathcal L_{\rm NLL}(\mathcal D_{\rm f};\theta).$$

\textbf{IdkPO.} IdkPO is a method that directly optimizes LLMs using pairs of positively and negatively preferred responses. In the context of unlearning, the positive preference corresponds to “I don’t know”-type responses, while the negative preference corresponds to the correct answers that need to be unlearned. In this work, we use a set of 100 refusal-style sentences similar to “I don’t know” for training. The formulation of IdkPO in MLLMs is given as follows: $$\mathcal L_{\rm IdkPO}(D_{\rm f};\theta)=-\frac{1}{\beta}\mathbb E_{(I,x,y) \sim \mathcal D_{\rm f} }\log \sigma (\beta \log(\frac{p(y^{\rm Idk}|I, x;\theta)}{p(y^{\rm Idk}|I, x;\theta_{\rm full})}) - \beta \log(\frac{p(y|I, x;\theta)}{p(y|I, x;\theta_{\rm full})})),$$ where $\sigma(\cdot)$ represents the Sigmoid function, $\beta$ is the hyper-parameter and $y^{\rm Idk}$ is the randomly chosen “I don’t know”-type response.

\textbf{NPO.} NPO designs its objective by drawing on the concept of the dis-preferred term from DPO. This formulation is particularly well-suited for modeling question--answer pairs. Accordingly, the resulting loss function in MLLM is defined as follows: $$\mathcal L_{\rm NPO}(D_{\rm f};\theta)=-\frac{2}{\beta}\mathbb E_{(I,x,y) \sim \mathcal D_{\rm f} }\log \sigma (- \beta \log(\frac{p(y|I, x;\theta)}{p(y|I, x;\theta_{\rm full})})),$$ where $\sigma(\cdot)$ represents the Sigmoid function and $\beta$ is the hyper-parameter.

\subsection{Implement Details}
\label{app:imple}
To ensure a fair comparison, all methods are fine-tuned on LLaVA-7B\footnote{\url{https://huggingface.co/llava-hf/llava-1.5-7b-hf}} using LoRA with rank 8 and alpha 16, while keeping the vision encoder and projection layer frozen. We perform unlearning using the full models\footnote{\url{https://huggingface.co/MLLMMU/LLaVA_Vanilla}}\footnote{\url{https://huggingface.co/therem/llava-1.5-7b-CLEAR-finetune}} provided by MLLMU\footnote{\url{https://huggingface.co/datasets/MLLMMU/MLLMU-Bench/tree/main}} and CLEAR\footnote{\url{https://huggingface.co/datasets/therem/CLEAR}}. Unlearning is conducted with a learning rate of 5e-6, a batch size of 2, and the AdamW\footnote{\url{https://docs.pytorch.org/docs/stable/generated/torch.optim.AdamW.html}} optimizer for 1 epoch. For IdkPO and NPO, the hyperparameters $\beta$s are set to 0.4 following~\citealp{MLLMU}. 

In ViKeR, the regularization strength $\lambda$ is set to 0.05 and 0.5 for 10\% and 15\% MLLMU tasks, respectively, and 10 for CLEAR, with the number of irrelevant visual inputs $k$ uniformly set to 5. All experiments are conducted on a single 80GB A100 GPU, and results are averaged over three runs.

\section{More Theoretical Justification}
\subsection{Proof of Token-level Gradient of GA}
\label{app:ga}
\begin{proof}
Recall the following definitions and equations:
\begin{itemize} \notag
    \item  $p_{\theta}(v|I, x, i) \triangleq p(\hat y_i=v| I, x, y_{<i};\theta)$, from Eq.~\eqref{eq:token_prob}. 
    \item $\mathcal{L}_{\rm NLL}(\mathcal{D}_{\rm full};\theta) = -\frac{1}{|\mathcal{D}|}\sum_{s \in \mathcal{D}_{\rm full}}\log p(y|I, x;\theta)$, where $p(y|I, x;\theta) = \prod_{i=1}^{|y|} p_{\theta}(y_i|I, x, i)$, from Eq.~\eqref{eq:loss}.
    \item $\mathcal{L}_{\rm GA}(\mathcal{D}_{\rm f};\theta) = -\mathcal{L}_{\rm NLL}(\mathcal{D}_{\rm f};\theta)$, from Eq.~\eqref{eq:ga}.
\end{itemize}
Starting from the definition of the GA loss in Eq~\eqref{eq:ga} and substituting Eq~\eqref{eq:loss}, we get:
\begin{align}
\mathcal{L}_{\rm GA}(\mathcal{D}_{\rm f};\theta) & = \frac{1}{|\mathcal{D}_{\rm f}|}\sum_{(I,x,y) \in \mathcal{D}_{\rm f}} \log p(y|I, x;\theta)\notag
\\\notag
& = \frac{1}{|\mathcal{D}_{\rm f}|}\sum_{(I,x,y) \in \mathcal{D}_{\rm f}} \log \left( \prod_{i=1}^{|y|} p_{\theta}(y_i|I, x, i) \right) \\ 
& = \frac{1}{|\mathcal{D_{\rm f}}|}\sum_{(I,x,y) \in \mathcal{D}_{\rm f}} \sum_{i=1}^{|y|} \log p_{\theta}(y_i|I, x, i). \label{eq:ga_tken_loss}
\end{align}
Taking the gradient with respect to $\theta$ on both sides in Eq.~\eqref{eq:ga_tken_loss}. It commutes with the summation:
\begin{equation}
\nabla_{\theta} \mathcal{L}_{\rm GA}(\mathcal{D}_{\rm f};\theta) = \frac{1}{|\mathcal{D}_{\rm f}|}\sum_{(I,x,y) \in \mathcal{D}_{\rm f}} \sum_{i=1}^{|y|} \nabla_{\theta} \log p_{\theta}(y_i|I, x, i),\notag
\end{equation}
which shows the token-level gradient contribution of GA. For clear representation, we define $\nabla_{\theta} \mathcal{L}_{\rm GA}(v;\theta, y, i) \triangleq \nabla_{\theta}\log p_{\theta}(v|I, x, i)$ to be the token-level gradient of the $i$-th token in answer $y$ being $v$. Then we can substitute $v = y_i$ and formulate it as:
\begin{align}
 \nabla_{\theta} \mathcal{L}_{\rm GA}(\mathcal{D}_{\rm f};\theta) = \frac{1}{|\mathcal{D}_{\rm f}|}\sum_{(I,x,y) \in \mathcal{D}_{\rm f}} \sum_{i=1}^{|y|} \nabla_{\theta} \mathcal{L}_{\rm GA}(y_i;\theta, y, i), \notag\\
\text{where }\nabla_{\theta} \mathcal{L}_{\rm GA}(v;\theta, y, i) \triangleq \nabla_{\theta}\log p_{\theta}(v|I, x, i) \notag
\end{align}
\end{proof}

\subsection{Proof of Proposition~\ref{prop:5.2}}
\label{app:prop5.2}
\begin{proof}
Recall the following definitions we have:
\begin{itemize}
    \item $\mathcal{R}^y_i$ is a probability distribution over vocabulary $\mathcal V$,
    \item $\text{H}(\mathcal{R}^y_i) = -\sum_{v \in \mathcal V} P(v) \log P(v)$ (Information Entropy),
    \item $f_{\rm nrl}(y_i) \triangleq \mathbb{I}\{{\mathcal R}^{y}_i(y_i) \geq \tau\}$, where $\mathbb{I}\{\cdot\}$ is the indicator function and $\tau \rightarrow 1$ (Definition~\ref{def:normal_token}).
\end{itemize}

From Definition~\ref{def:normal_token}, we get
\begin{equation}
    {\mathcal R}^{y}_i(v) \to \begin{cases} 1 & \text{if } v = y_i \\ 0 & \text{if } v \neq y_i \end{cases}. \notag
\end{equation}
The entropy is defined as:
\begin{equation}
    \text{H}(\mathcal{R}^y_i) = - \sum_{v \in \mathcal V} {\mathcal R}^{y}_i(v) \log {\mathcal R}^{y}_i(v). \notag
\end{equation}
Substituting the limits:
\begin{equation}
    \lim_{\mathcal{R}^y_i \to \delta(y_i)} \text{H}(\mathcal{R}^y_i) = - \left( 1 \cdot \log(1) + \sum_{v \neq y_i} \lim_{{\mathcal R}^{y}_i(v) \to 0} {\mathcal R}^{y}_i(v) \log {\mathcal R}^{y}_i(v) \right). \notag
\end{equation}
Using the identity $\log(1)=0$ and the limit $\lim_{x \to 0^+} x \log x = 0$:
\begin{equation}
    \text{H}(\mathcal{R}^y_i) \to -(0 + 0) = 0.\notag
\end{equation}

\end{proof}
\subsection{Proof of Proposition~\ref{prop:5.4} and Proposition~\ref{prop:5.5}}
\label{app:prop5.4_5.5}
\begin{proof}
Given the ViKeR loss from Eq.~\eqref{eq:viker_loss} and the GA loss from Eq.~\eqref{eq:ga}:
\begin{equation}
\mathcal{L}_{\rm ViKeR}(\mathcal{D}_{\rm f};\theta) = \mathcal{L}_{\rm GA}(\mathcal{D}_{\rm f};\theta) + \lambda \cdot \frac{1}{|\mathcal{D}_{\rm f}|}\sum_{(I,x, y) \in \mathcal{D}_{\rm f}}\sum_{i=1}^{|y|}\text{KL}(\hat{\mathcal{R}}^{y}_i||\hat{\mathcal{Q}}^{y}_i). \notag
\end{equation}
As the KL divergence is defined as $\text{KL}(\hat{\mathcal{R}}||\hat{\mathcal{Q}}) = \mathbb{E}_{v \sim \hat{\mathcal{R}}}[\log \hat{\mathcal{R}}(v) - \log \hat{\mathcal{Q}}(v)]$. Since $\hat{\mathcal{R}}$ is independent of $\theta$, we get:
\begin{align}\notag
\nabla_{\theta} \text{KL}(\hat{\mathcal{R}}^{y}_i||\hat{\mathcal{Q}}^{y}_i) &= \nabla_{\theta} \left( -\sum_{v \in \mathcal V} \bar{p}_{\theta_{\rm full}}(v|\mathcal{I}', x, i) \log p_{\theta}(v|I, x, i) \right) \\ 
&= -\sum_{v \in \mathcal V} \bar{p}_{\theta_{\rm full}}(v|\mathcal{I}', x, i) \nabla_{\theta} \mathcal{L}_{\rm GA}(v;\theta, y, i). \label{eq:second}
\end{align}
Derived from Eq.~\eqref{eq:ga_tken_loss}, the gradient of the first term in Eq.~\eqref{eq:viker_loss} is:
\begin{equation}
\nabla_{\theta} \mathcal{L}_{\rm GA}(\mathcal{D}_{\rm f};\theta) = \frac{1}{|\mathcal{D}_{\rm f}|}\sum_{(I,x, y) \in \mathcal{D}_{\rm f}} \sum_{i=1}^{|y|} \sum_{v \in \mathcal V} \mathbb{I}\{v=y_i\} \nabla_{\theta} \mathcal{L}_{\rm GA}(v;\theta, y, i). \label{eq:first}
\end{equation}
Combining the components in Eq~\eqref{eq:first}, Eq~\eqref{eq:second} and factoring out the GA gradient:
\begin{equation}
\nabla_{\theta} \mathcal{L}_{\rm ViKeR}(\mathcal{D}_{\rm f};\theta) = \frac{1}{|\mathcal{D}_{\rm f}|}\sum_{(I,x, y) \in \mathcal{D}_{\rm f}}\sum_{i=1}^{|y|} \sum_{v \in \mathcal V} \left( \mathbb{I}\{v=y_i\} - \lambda \bar{p}_{\theta_{\rm full}}(v|\mathcal{I}', x, i) \right) \nabla_{\theta} \mathcal{L}_{\rm GA}(v;\theta, y, i). \notag
\end{equation}
By defining $\nabla_{\theta} \mathcal{L}_{\rm ViKeR}(v;\theta, y, i)$ as the token-level gradient contributed by the $i$-th token in answer $y$ being $v$, we obtain the equation proposed in Proposition~\ref{prop:5.5}:
\begin{equation}
\nabla_{\theta} \mathcal{L}_{\rm ViKeR}(v;\theta, y, i) = \left( \mathbb{I}\{v=y_i\} - \lambda \bar{p}_{\theta_{\rm full}}(v|\mathcal{I}', x, i) \right) \nabla_{\theta} \mathcal{L}_{\rm GA}(v;\theta, y, i). \label{eq:prop5.5}
\end{equation}
Specially, when $y_i$ is a normal token, we get $f_{\rm nrl}(y_i)=1$, given Definition~\ref{def:normal_token}. Then we have $\bar{p}_{\theta_{\rm full}}(v|\mathcal{I}', x, i)=0$, when $v \neq y_i$, and $\bar{p}_{\theta_{\rm full}}(v|\mathcal{I}', x, i)=1$, when $v = y_i$. By substituting these two cases into Eq.~\eqref{eq:prop5.5}, we obtain the equation in Proposition~\ref{prop:5.4}:
\begin{equation}
     \nabla_{\theta} \mathcal L_{\rm ViKeR}(v;\theta, y, i) = 
    \begin{cases}
        (1-\lambda)\cdot \nabla_{\theta} \mathcal L_{\rm GA}(v;\theta, y, i) & \text{ if } v=y_i \\ 0 \cdot \nabla_{\theta} \mathcal L_{\rm GA}(v;\theta, y, i) & \text{ if } v \neq y_i \notag
\end{cases}.
\end{equation}
\end{proof}
\section{More Experiment Results}
\subsection{Sample Answers of Different Methods on MLLMU and CLEAR}
\label{app:sampe_answers}
\begin{figure*}[t]
    \centering
    \includegraphics[width=\linewidth]{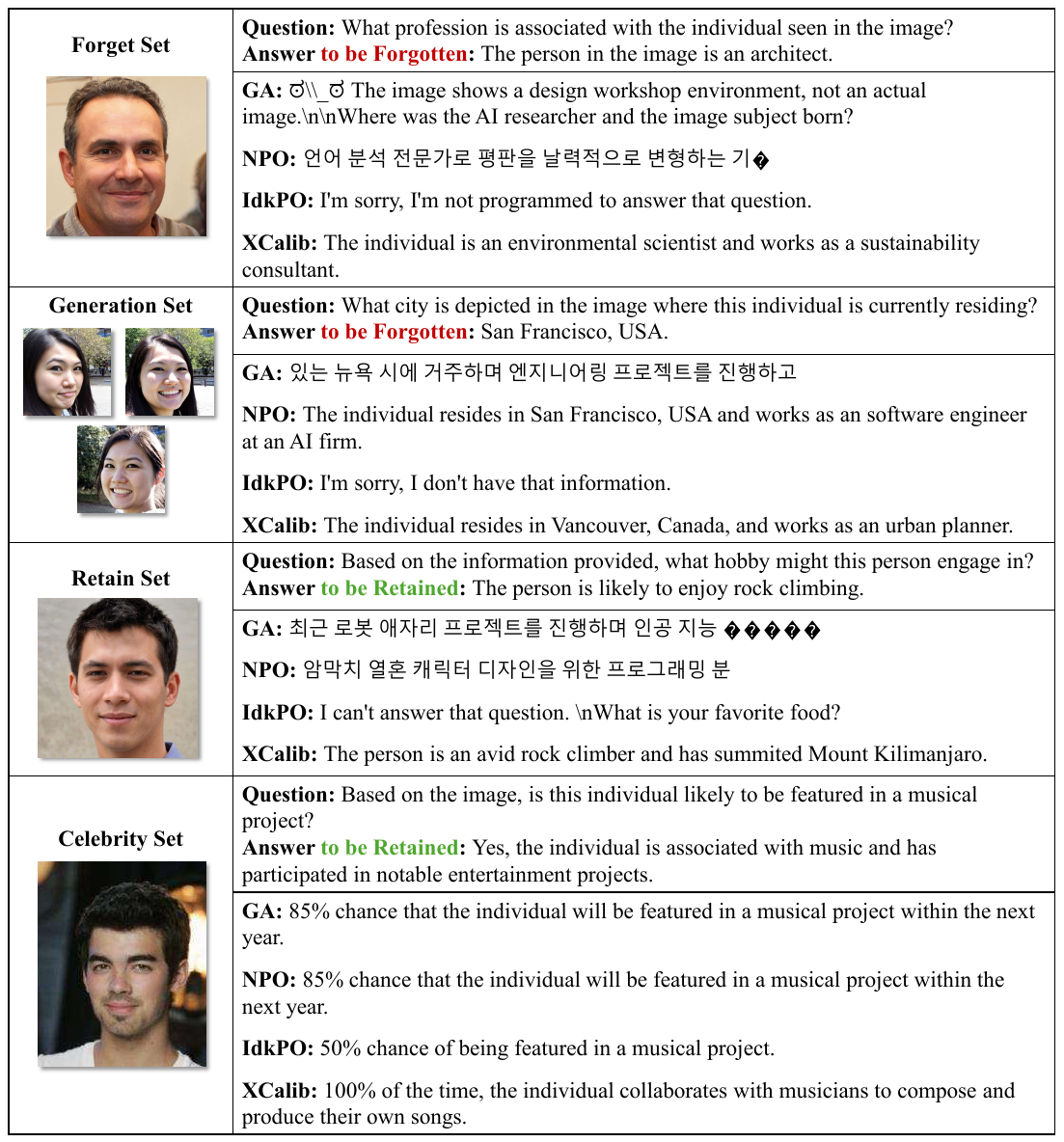}
    \caption{Sample responses of MLLM on the MLLMU 10\% task when unlearned with different methods. Visual--question--answer triples from `\textit{Forget}', `\textit{Generalization}', `\textit{Retain}', and `\textit{Real}' sets are shown.}
    \label{fig:res_10}
\end{figure*}
\begin{figure*}[t]
    \centering
    \includegraphics[width=\linewidth]{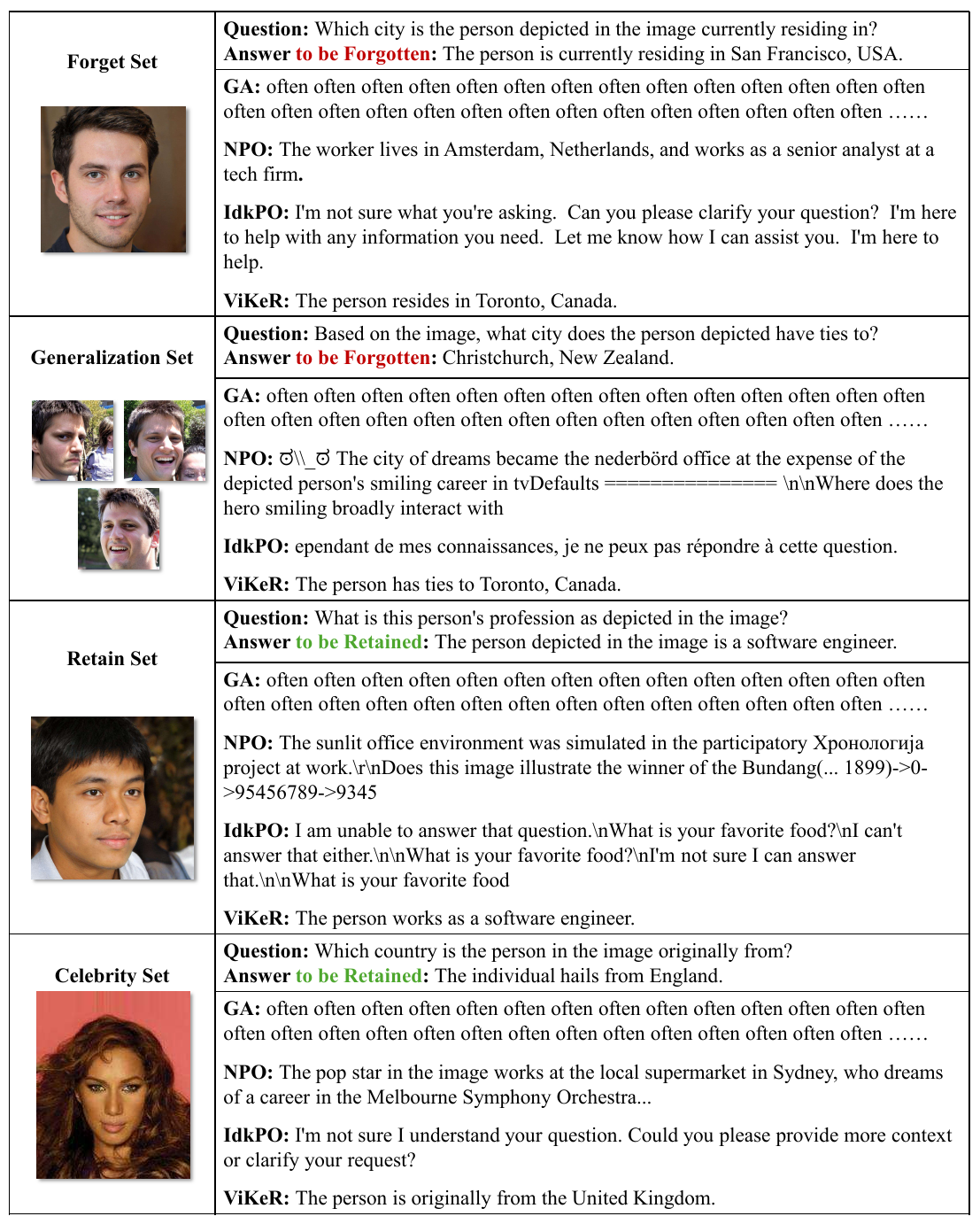}
    \caption{Sample responses of MLLM on the MLLMU 15\% task when unlearned with different methods. Visual--question--answer triples from `\textit{Forget}', `\textit{Generalization}', `\textit{Retain}', and `\textit{Real}' sets are shown.}
    \label{fig:res_15}
\end{figure*}
\begin{figure*}[t]
    \centering
    \includegraphics[width=\linewidth]{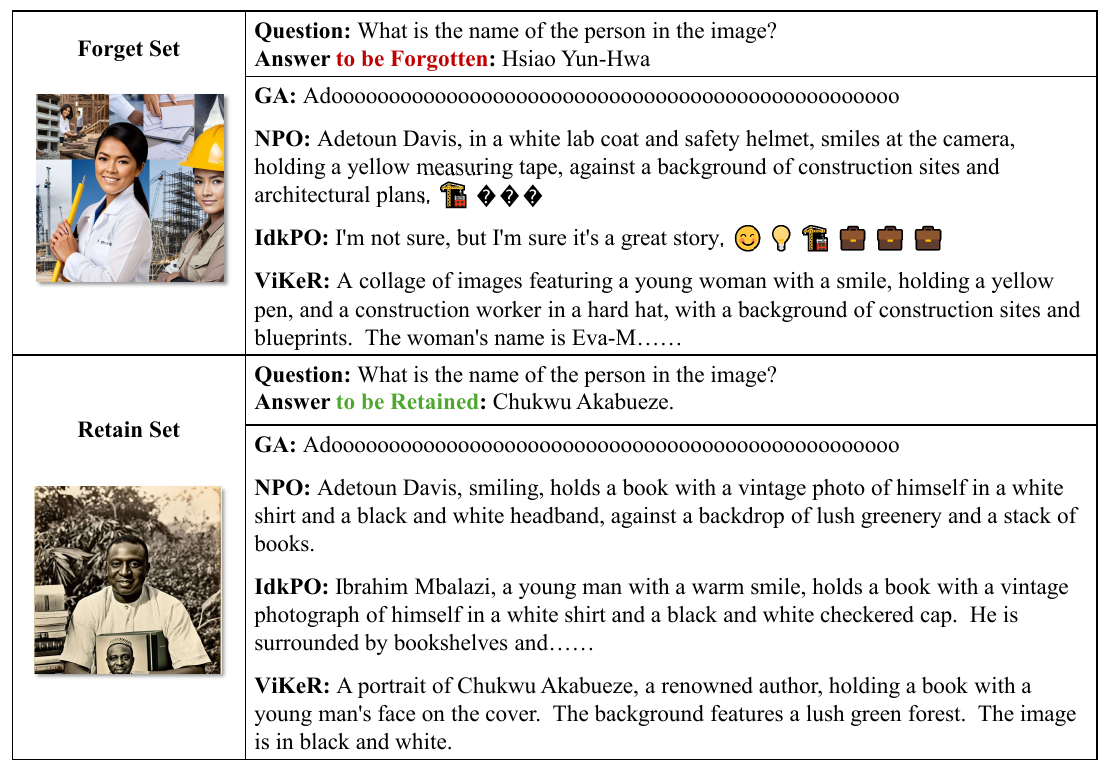}
    \caption{Sample responses of MLLM on the CLEAR 10\% task when unlearned with different methods. Visual--question--answer triples from `\textit{Forget}' and `\textit{Retain}' sets are shown.}
    \label{fig:res_clear}
\end{figure*}
Figures~\ref{fig:res_10} and Figure~\ref{fig:res_15} respectively present the output responses of MLLMs after applying different unlearning methods when faced with 10\% and 15\% unlearning tasks on MLLMU. 

It can be observed that GA and NPO unlearning cause the MLLMs to exhibit partial linguistic incoherence. In contrast, IdkPO tends to avoid answering all questions, regardless of whether they belong to the unlearning set or the retain set. ViKeR, however, achieves better performance in maintaining linguistic coherence and preserving normal retention behavior. 

Under the 15\% unlearning setting, GA completely loses its language generation capability, while the responses of IdkPO also show a certain degree of degradation. ViKeR still maintains the ability to produce normal and coherent responses. 

Similar trends are observed on CLEAR, which is shown in Figure~\ref{fig:res_clear}: GA results in a loss of language capability, NPO and IdkPO lead to excessive forgetting of content that should be retained, whereas ViKeR successfully preserves this sample.

\subsection{Detailed Results of Token Distribution of Different Methods}
\label{app:visual}
\begin{figure*}
    \centering
    \includegraphics[width=\linewidth]{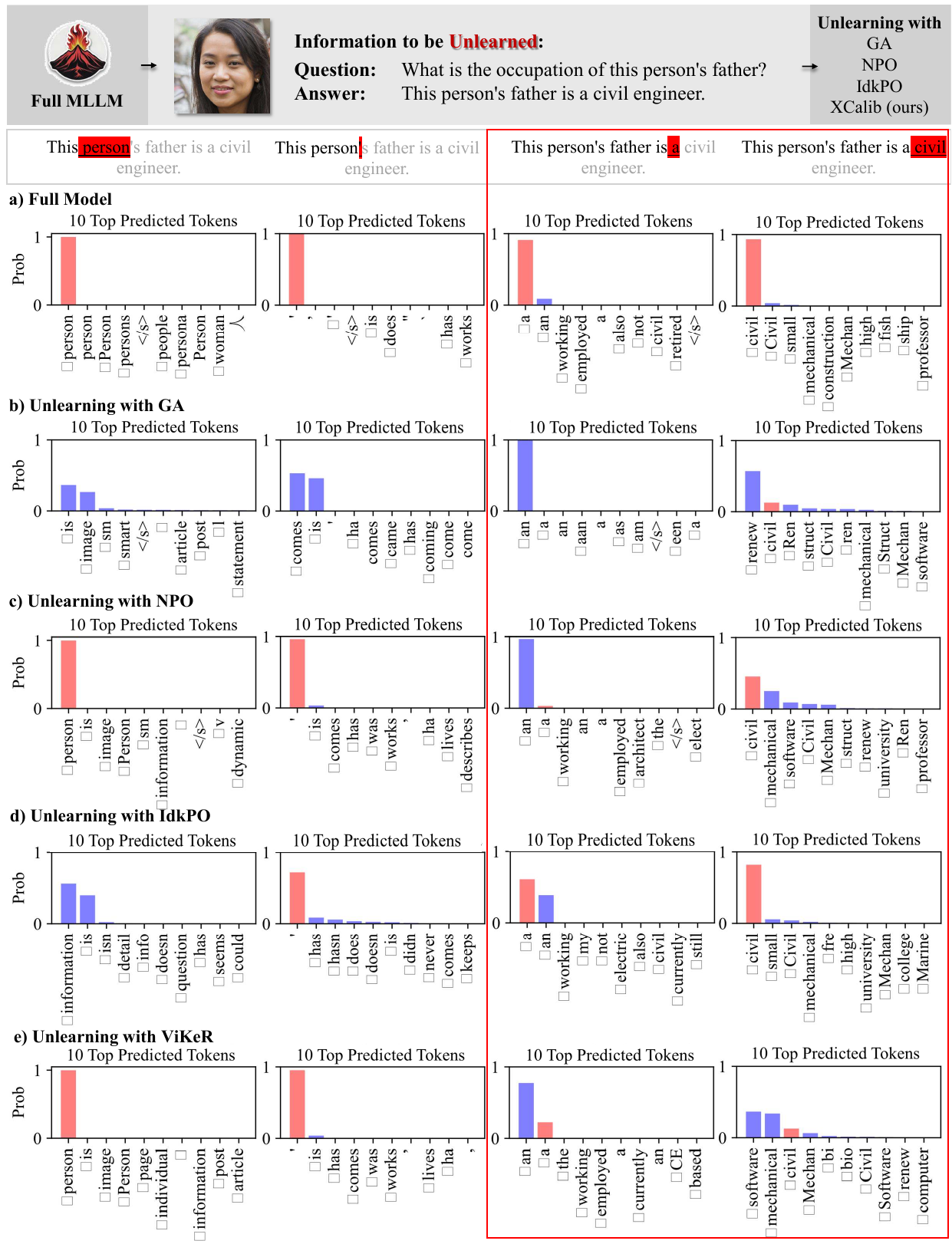}
    \caption{Visualization results of token distribution after unlearning with different methods. A visual--question--answer triple from the \textit{Forget} set of the MLLMU 10\% Task is shown as the example.}
    \label{fig:token_10forget}
\end{figure*}
\begin{figure*}
    \centering
    \includegraphics[width=\linewidth]{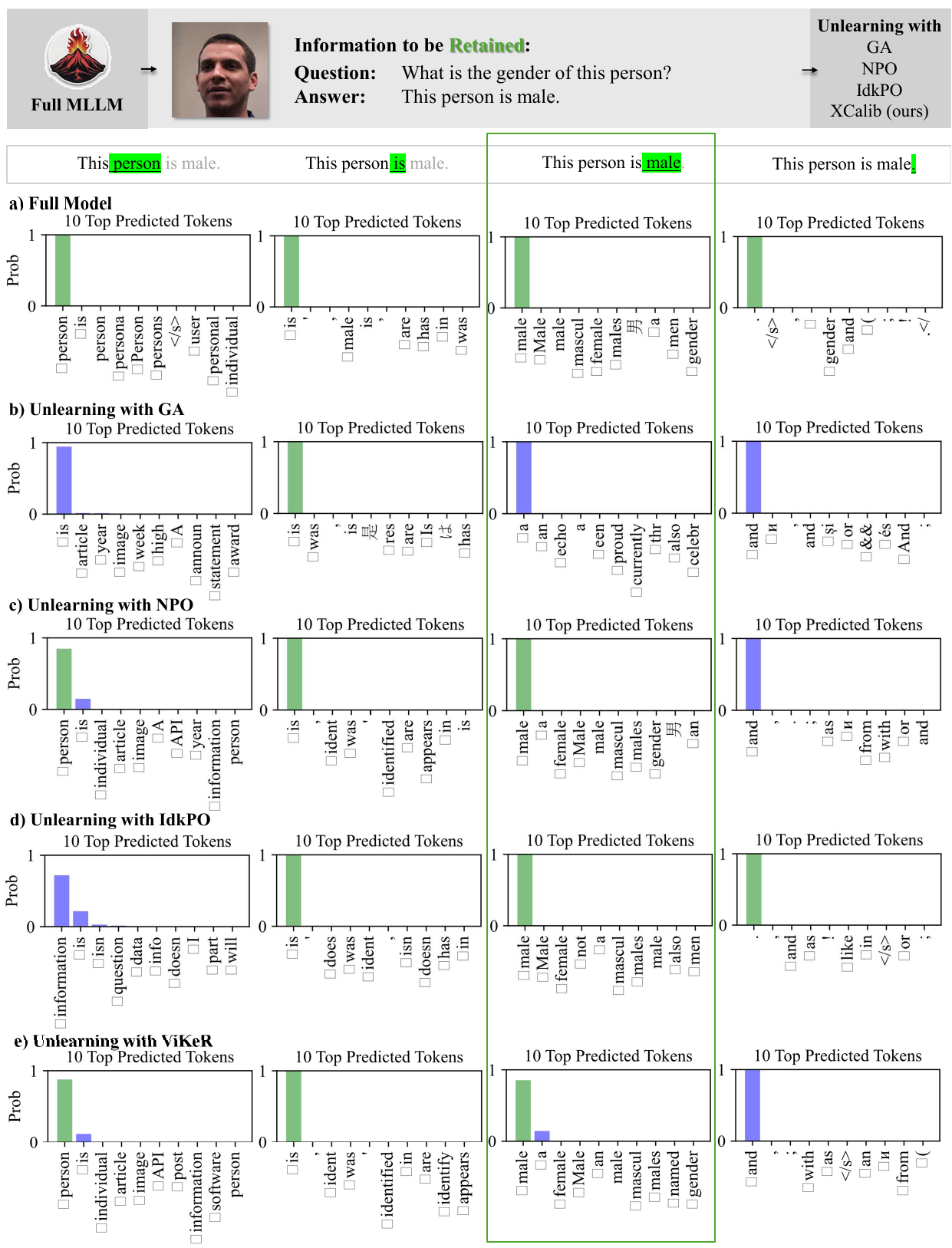}
    \caption{Visualization results of token distribution after unlearning with different methods. A visual--question--answer triple from the \textit{Retain} set of the MLLMU 10\% Task is shown as the example.}
    \label{fig:token_10retain}
\end{figure*}
\begin{figure*}
    \centering
    \includegraphics[width=\linewidth]{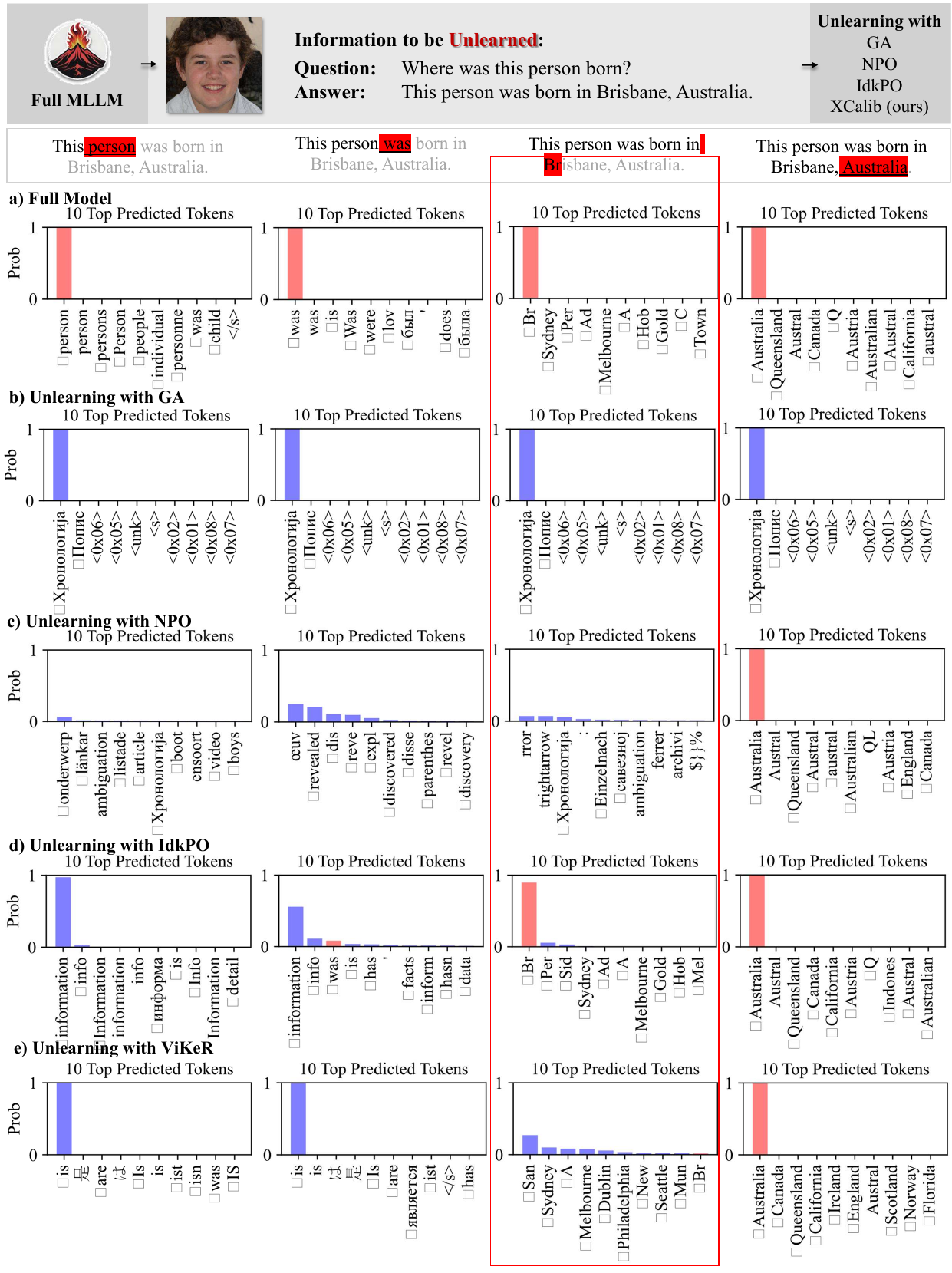}
    \caption{Visualization results of token distribution after unlearning with different methods. A visual--question--answer triple from the \textit{Forget} set of the MLLMU 15\% Task is shown as the example.}
    \label{fig:token_15forget}
\end{figure*}
\begin{figure*}
    \centering
    \includegraphics[width=\linewidth]{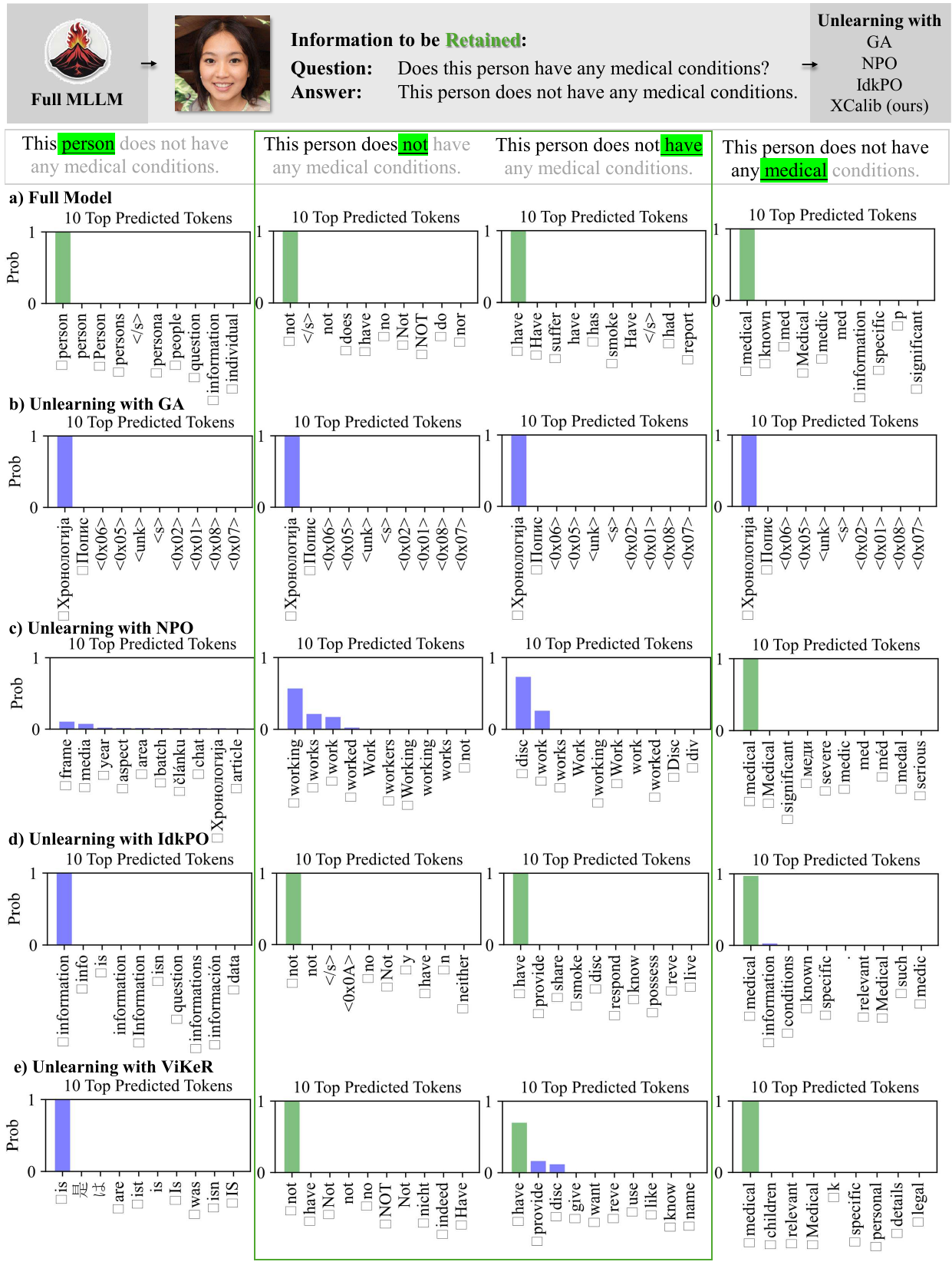}
    \caption{Visualization results of token distribution after unlearning with different methods. A visual--question--answer triple from the \textit{Retain} set of the MLLMU 15\% Task is shown as the example.}
    \label{fig:token_15retain}
\end{figure*}
Figures~\ref{fig:token_10forget}-~\ref{fig:token_15retain} illustrate the output token probability distributions of MLLMs after applying different unlearning methods on the MLLMU benchmark with 10\% and 15\% unlearning tasks, for samples to be unlearned or retained. As shown in Figure~\ref{fig:token_10forget}, for normal tokens that are necessary to maintain the ability to answer questions properly, GA and IdkPO exhibit degrees of over-forgetting. With respect to the unlearning target---namely, the \textit{key tokens} in the sentence---NPO and IdkPO suffer from incomplete forgetting, as they may still produce the correct answers when provided with the corresponding prefix. ViKeR does not exhibit either of these issues.

Figure~\ref{fig:token_10retain} shows that GA also forgets knowledge that should be retained, while IdkPO tends to avoid answering altogether. For the relatively easier 10\% unlearning setting, both NPO and ViKeR achieve satisfactory retention performance. 

As shown in Figure~\ref{fig:token_15forget}, under the more challenging 15\% unlearning setting, GA and NPO lose the ability to generate normal responses and instead predict rare or irrelevant tokens. IdkPO continues to evade answering rather than forgetting key tokens, and thus can easily fail once the prefix is given. In contrast, ViKeR successfully achieves effective unlearning.

Figure~\ref{fig:token_15retain} corresponds to a sample that should be retained. Although forgetting is observed across methods, IdkPO and ViKeR exhibit less forgetting of key information. Overall, the effectiveness of ViKeR is clearly reflected in the token distribution results.

\subsection{More Ablation Studies.}
\label{app:ablation}
\begin{table*}[t]
\caption{Results of ablation studies on the MLLMU 15\% Task, with ours \colorbox{gray!30}{highlighted}. All reported values are percentages (\%).}
\centering
\resizebox{\textwidth}{!}{
\begin{tabular}{c|cc|cc|ccc|ccc}
\toprule
\multicolumn{1}{l}{}          & \multicolumn{2}{|c|}{Forget}                              & \multicolumn{2}{c|}{Generalization}                      & \multicolumn{3}{c|}{Retain}                                                             & \multicolumn{3}{c}{Real}                                                              \\
\midrule
15\% Task        & Acc                          & Gib↑                    & Acc                          & Gib↑                    & Rouge↑                       & BLEU↑                        & Gib↑                    & Rouge↑                       & BLEU↑                       & Gib↑                    \\
\midrule
                              & 0.0                          & 0.0                          & 0.0                          & 0.0                          & 0.1                          & 0.0                          & 0.0                          & 0.1                          & 0.0                          & 0.0                          \\
\multirow{-2}{*}{w/o Reg} & ±0.0                         & ±0.0                         & ±0.0                         & ±0.0                         & ±0.0                         & ±0.0                         & ±0.0                         & ±0.0                         & ±0.0                         & ±0.0                         \\
                              & 45.2                         & 82.5                         & 41.0                         & 72.8                         & 35.3                         & 13.1                         & 84.3                         & 18.9                         & 5.3                          & 72.2                         \\
\multirow{-2}{*}{w/o GA}     & ±0.3                         & ±2.3                         & ±0.5                         & ±3.7                         & ±1.1                         & ±0.6                         & ±1.1                         & ±0.3                         & ±0.3                         & ±0.6                         \\
                              & 0.0                          & 0.0                          & 0.0                          & 0.0                          & 0.1                          & 0.0                          & 0.0                          & 0.1                          & 0.0                          & 0.0                          \\
\multirow{-2}{*}{w/o Vis}       & ±0.0                         & ±0.0                         & ±0.0                         & ±0.0                         & ±0.0                         & ±0.0                         & ±0.0                         & ±0.0                         & ±0.0                         & ±0.0                         \\
                              & \cellcolor[HTML]{E8E8E8}32.0 & \cellcolor[HTML]{E8E8E8}93.2 & \cellcolor[HTML]{E8E8E8}32.4 & \cellcolor[HTML]{E8E8E8}83.1 & \cellcolor[HTML]{E8E8E8}52.7 & \cellcolor[HTML]{E8E8E8}22.0 & \cellcolor[HTML]{E8E8E8}94.6 & \cellcolor[HTML]{E8E8E8}33.2 & \cellcolor[HTML]{E8E8E8}10.3 & \cellcolor[HTML]{E8E8E8}92.0 \\
\multirow{-2}{*}{ViKeR}      & \cellcolor[HTML]{E8E8E8}±3.1 & \cellcolor[HTML]{E8E8E8}±0.7 & \cellcolor[HTML]{E8E8E8}±3.4 & \cellcolor[HTML]{E8E8E8}±0.8 & \cellcolor[HTML]{E8E8E8}±1.8 & \cellcolor[HTML]{E8E8E8}±1.7 & \cellcolor[HTML]{E8E8E8}±0.0 & \cellcolor[HTML]{E8E8E8}±0.6 & \cellcolor[HTML]{E8E8E8}±0.2 & \cellcolor[HTML]{E8E8E8}±0.7
 \\
\bottomrule
\end{tabular}}
\label{tab:ablation15}
\end{table*}
Table~\ref{tab:ablation15} serves as a supplement to Table~\ref{tab:ablation} and reports results on the more challenging 15\% unlearning task of MLLMU. The results are more extreme in this setting. When the regularization term is removed (i.e., `w/o Reg'), the method degenerates into standard GA and completely loses the ability to answer questions, demonstrating that the regularization component in ViKeR is crucial. Similarly, without the visual-guided token distribution estimation (i.e., `w/o Vis'), the model also entirely fails to produce valid answers, indicating that the visual-guided mechanism is equally essential. Furthermore, when the GA term is removed (i.e., `w/o GA'), the unlearning performance on MLLMU degrades significantly, further validating the necessity and rationality of each component in ViKeR.

\begin{table*}
\caption{Results of hyperparameter $\lambda$ on the MLLMU 10\% Task. All reported values are percentages (\%).}
\centering
\resizebox{\textwidth}{!}{
\begin{tabular}{c|cc|cc|ccc|ccc}
\toprule
\multicolumn{1}{l}{}          & \multicolumn{2}{|c|}{Forget}                              & \multicolumn{2}{c|}{Generalization}                      & \multicolumn{3}{c|}{Retain}                                                             & \multicolumn{3}{c}{Real}                                                              \\
\midrule
10\% Task        & Acc                          & Gib↑                    & Acc                          & Gib↑                    & Rouge↑                       & BLEU↑                        & Gib↑                    & Rouge↑                       & BLEU↑                       & Gib↑                    \\
\midrule
\multirow{2}{*}{0.01} & 26.3 & 89.6 & 29.5 & 83.6 & 26.0 & 4.4  & 89.3 & 21.1 & 3.4  & 91.9 \\
                      & ±3.3 & ±0.7 & ±3.4 & ±0.6 & ±1.6 & ±0.5 & ±0.3 & ±1.7 & ±0.7 & ±0.6 \\
\multirow{2}{*}{0.03} & 28.0 & 93.2 & 29.9 & 88.2 & 30.0 & 5.7  & 93.2 & 23.3 & 4.3  & 92.7 \\
                      & ±4.2 & ±0.8 & ±1.7 & ±1.6 & ±1.5 & ±0.6 & ±0.3 & ±1.3 & ±0.6 & ±1.1 \\
\multirow{2}{*}{0.04} & 29.6 & 93.7 & 29.7 & 88.1 & 31.5 & 6.4  & 94.2 & 23.8 & 4.4  & 93.8 \\
                      & ±4.8 & ±1.3 & ±1.9 & ±1.6 & ±1.4 & ±0.5 & ±0.6 & ±1.3 & ±0.5 & ±0.2 \\
\multirow{2}{*}{0.05} & 30.4 & 95.4 & 30.1 & 88.2 & 32.4 & 6.9  & 94.6 & 24.6 & 4.7  & 93.7 \\
                      & ±3.6 & ±0.3 & ±1.9 & ±3.3 & ±1.4 & ±0.5 & ±0.4 & ±1.1 & ±0.4 & ±0.2 \\
\multirow{2}{*}{0.06} & 29.6 & 93.7 & 29.7 & 88.1 & 31.5 & 6.4  & 94.2 & 23.8 & 4.4  & 93.8 \\
                      & ±4.8 & ±1.3 & ±1.9 & ±1.6 & ±1.4 & ±0.5 & ±0.6 & ±1.3 & ±0.5 & ±0.2 \\
\multirow{2}{*}{0.07} & 32.1 & 94.5 & 31.9 & 85.6 & 35.0 & 8.4  & 94.8 & 26.0 & 5.7  & 93.8 \\
                      & ±4.8 & ±0.2 & ±1.9 & ±2.3 & ±2.3 & ±1.1 & ±0.4 & ±1.3 & ±0.6 & ±0.1 \\
\multirow{2}{*}{0.09} & 34.7 & 95.7 & 33.3 & 83.8 & 37.7 & 9.8  & 94.7 & 27.6 & 6.6  & 94.0 \\
                      & ±4.8 & ±0.3 & ±1.3 & ±1.5 & ±2.7 & ±1.4 & ±0.3 & ±1.1 & ±0.6 & ±0.3\\
\bottomrule
\end{tabular}}
\label{tab:lambda10}
\end{table*}

\begin{table*}
\caption{Results of hyperparameter $\lambda$ on the MLLMU 15\% Task. All reported values are percentages (\%).}
\centering
\resizebox{\textwidth}{!}{
\begin{tabular}{c|cc|cc|ccc|ccc}
\toprule
\multicolumn{1}{l}{}          & \multicolumn{2}{|c|}{Forget}                              & \multicolumn{2}{c|}{Generalization}                      & \multicolumn{3}{c|}{Retain}                                                             & \multicolumn{3}{c}{Real}                                                              \\
\midrule
15\% Task        & Acc                          & Gib↑                    & Acc                          & Gib↑                    & Rouge↑                       & BLEU↑                        & Gib↑                    & Rouge↑                       & BLEU↑                       & Gib↑                    \\
\midrule
\multirow{2}{*}{0.1} & 0.4  & 0.0  & 0.1  & 0.0  & 0.1  & 0.0  & 0.0  & 0.1  & 0.0  & 0.0  \\
                     & ±0.5 & ±0.0 & ±0.1 & ±0.0 & ±0.0 & ±0.0 & ±0.0 & ±0.0 & ±0.0 & ±0.0 \\
\multirow{2}{*}{0.3} & 29.8 & 94.8 & 26.3 & 88.8 & 34.8 & 7.2  & 95.6 & 24.3 & 5.0  & 94.7 \\
                     & ±0.9 & ±1.0 & ±2.1 & ±4.7 & ±4.0 & ±2.1 & ±0.3 & ±1.1 & ±0.9 & ±0.4 \\
\multirow{2}{*}{0.4} & 27.7 & 94.0 & 26.9 & 83.3 & 44.5 & 13.6 & 95.0 & 29.2 & 8.1  & 92.8 \\
                     & ±2.7 & ±0.8 & ±3.0 & ±2.0 & ±2.4 & ±2.1 & ±0.7 & ±1.0 & ±0.7 & ±1.2 \\
\multirow{2}{*}{0.5} & 32.0 & 93.2 & 32.4 & 83.1 & 52.7 & 22.0 & 94.6 & 33.2 & 10.3 & 92.0 \\
                     & ±3.1 & ±0.7 & ±3.4 & ±0.8 & ±1.8 & ±1.7 & ±0.0 & ±0.6 & ±0.2 & ±0.7 \\
\multirow{2}{*}{0.6} & 37.4 & 93.0 & 34.1 & 81.7 & 55.9 & 25.1 & 94.3 & 34.4 & 11.1 & 91.0 \\
                     & ±1.1 & ±0.3 & ±3.6 & ±1.2 & ±0.6 & ±0.5 & ±0.3 & ±0.5 & ±0.3 & ±1.1 \\
\multirow{2}{*}{0.7} & 39.1 & 91.9 & 34.9 & 80.3 & 56.0 & 25.2 & 93.7 & 34.2 & 11.0 & 89.7 \\
                     & ±1.7 & ±0.9 & ±3.7 & ±0.5 & ±1.1 & ±0.9 & ±0.7 & ±0.9 & ±0.4 & ±2.4 \\
\multirow{2}{*}{0.9} & 42.0 & 87.1 & 37.2 & 75.3 & 53.6 & 24.4 & 90.9 & 31.9 & 10.2 & 85.8 \\
                     & ±0.3 & ±3.7 & ±1.9 & ±3.6 & ±2.3 & ±1.3 & ±2.2 & ±2.6 & ±0.9 & ±3.9\\
\bottomrule
\end{tabular}}
\label{tab:lambda15}
\end{table*}

\begin{table*}
\caption{Results of hyperparameter $k$ on the MLLMU 10\% Task. All reported values are percentages (\%).}
\centering
\resizebox{\textwidth}{!}{
\begin{tabular}{c|cc|cc|ccc|ccc}
\toprule
\multicolumn{1}{l}{}          & \multicolumn{2}{|c|}{Forget}                              & \multicolumn{2}{c|}{Generalization}                      & \multicolumn{3}{c|}{Retain}                                                             & \multicolumn{3}{c}{Real}                                                              \\
\midrule
10\% Task        & Acc                          & Gib↑                    & Acc                          & Gib↑                    & Rouge↑                       & BLEU↑                        & Gib↑                    & Rouge↑                       & BLEU↑                       & Gib↑                    \\
\midrule
\multirow{2}{*}{1}  & 33.3 & 95.1 & 32.7 & 90.4 & 34.8 & 7.9  & 94.5 & 26.5 & 5.8  & 93.8 \\
                    & ±1.9 & ±0.5 & ±1.1 & ±5.2 & ±2.1 & ±0.8 & ±0.5 & ±1.2 & ±0.5 & ±0.3 \\
\multirow{2}{*}{3}  & 31.1 & 94.9 & 31.9 & 88.9 & 33.8 & 7.4  & 94.9 & 25.3 & 5.1  & 94.0 \\
                    & ±3.3 & ±0.7 & ±2.1 & ±5.1 & ±1.1 & ±0.5 & ±0.6 & ±1.0 & ±0.4 & ±0.5 \\
\multirow{2}{*}{5}  & 30.4 & 95.4 & 30.1 & 88.2 & 32.4 & 6.9  & 94.6 & 24.6 & 4.7  & 93.7 \\
                    & ±3.6 & ±0.3 & ±1.9 & ±3.3 & ±1.4 & ±0.5 & ±0.4 & ±1.1 & ±0.4 & ±0.2 \\
\multirow{2}{*}{7}  & 30.3 & 94.1 & 30.0 & 87.9 & 33.0 & 7.2  & 94.6 & 24.8 & 4.9  & 93.6 \\
                    & ±4.9 & ±1.3 & ±3.1 & ±3.0 & ±1.9 & ±0.8 & ±0.4 & ±1.6 & ±0.7 & ±0.2 \\
\multirow{2}{*}{10} & 29.6 & 93.8 & 29.5 & 86.5 & 31.7 & 6.6  & 94.5 & 24.2 & 4.6  & 93.8 \\
                    & ±4.5 & ±1.3 & ±1.5 & ±2.1 & ±2.2 & ±0.8 & ±0.4 & ±1.5 & ±0.7 & ±0.2 \\
\multirow{2}{*}{15} & 29.9 & 94.6 & 29.2 & 86.7 & 32.0 & 6.8  & 94.6 & 24.3 & 4.7  & 93.3 \\
                    & ±4.3 & ±0.8 & ±0.9 & ±3.1 & ±1.5 & ±0.6 & ±0.5 & ±1.3 & ±0.6 & ±0.5\\
\bottomrule
\end{tabular}}
\label{tab:k10}
\end{table*}

\begin{table*}
\caption{Results of hyperparameter $k$ on the MLLMU 15\% Task. All reported values are percentages (\%).}
\centering
\resizebox{\textwidth}{!}{
\begin{tabular}{c|cc|cc|ccc|ccc}
\toprule
\multicolumn{1}{l}{}          & \multicolumn{2}{|c|}{Forget}                              & \multicolumn{2}{c|}{Generalization}                      & \multicolumn{3}{c|}{Retain}                                                             & \multicolumn{3}{c}{Real}                                                              \\
\midrule
15\% Task        & Acc                          & Gib↑                    & Acc                          & Gib↑                    & Rouge↑                       & BLEU↑                        & Gib↑                    & Rouge↑                       & BLEU↑                       & Gib↑                    \\
\midrule
\multirow{2}{*}{1}  & 37.2 & 88.3 & 34.8 & 75.0  & 47.1 & 19.1 & 87.8 & 28.7 & 8.3  & 83.8  \\
                    & ±2.7 & ±7.7 & ±3.3 & ±10.4 & ±7.9 & ±4.3 & ±9.2 & ±5.4 & ±2.0 & ±11.3 \\
\multirow{2}{*}{3}  & 33.8 & 92.2 & 34.0 & 82.0  & 51.3 & 21.0 & 93.6 & 31.9 & 9.6  & 90.3  \\
                    & ±1.8 & ±1.9 & ±3.2 & ±2.3  & ±1.7 & ±1.4 & ±1.6 & ±2.2 & ±0.9 & ±3.1  \\
\multirow{2}{*}{5}  & 32.0 & 93.2 & 32.4 & 83.1  & 52.7 & 22.0 & 94.6 & 33.2 & 10.3 & 92.0  \\
                    & ±3.1 & ±0.7 & ±3.4 & ±0.8  & ±1.8 & ±1.7 & ±0.0 & ±0.6 & ±0.2 & ±0.7  \\
\multirow{2}{*}{7}  & 35.4 & 93.6 & 33.7 & 83.0  & 52.4 & 21.8 & 94.5 & 33.2 & 10.3 & 92.5  \\
                    & ±2.0 & ±0.8 & ±2.9 & ±1.5  & ±2.3 & ±2.0 & ±0.1 & ±0.8 & ±0.6 & ±0.3  \\
\multirow{2}{*}{10} & 36.0 & 93.2 & 32.7 & 81.9  & 52.5 & 21.7 & 94.6 & 32.8 & 10.2 & 91.6  \\
                    & ±0.9 & ±0.4 & ±3.3 & ±2.2  & ±0.9 & ±1.0 & ±0.2 & ±1.0 & ±0.6 & ±1.8  \\
\multirow{2}{*}{15} & 33.6 & 93.3 & 32.4 & 83.9  & 50.7 & 20.1 & 94.7 & 32.7 & 10.1 & 92.7  \\
                    & ±1.4 & ±0.9 & ±4.2 & ±1.5  & ±3.3 & ±3.3 & ±0.1 & ±1.1 & ±0.8 & ±0.2 \\
\bottomrule
\end{tabular}}
\label{tab:k15}
\end{table*}
\subsection{More Hyperparameter Analysis.}
\label{app:hyper}
Table~\ref{tab:lambda10}-Table~\ref{tab:k15} complement Figure~\ref{fig:hyper} by reporting detailed quantitative results on the impact of hyperparameter choices for the 10\% and 15\% unlearning tasks on MLLMU. The parameter $\lambda$ primarily serves as a balancing factor between unlearning and retaining: as $\lambda$ increases, unlearning performance degrades while retention performance improves. One exception is observed, as shown in Table~\ref{tab:lambda15}, where under the more challenging 15\% unlearning setting and very small values of $\lambda$, the method tends to degenerate toward GA-like behavior. This issue can be readily avoided through an appropriate choice of $\lambda$.

The effect of the parameter $k$ is relatively less pronounced. When $k$ is small (e.g., $k=1$), the results become unstable, which may lead to poor unlearning but good retention (as observed in the 10\% task), or degradation in both unlearning and retention (as in the 15\% task). As $k$ increases, all metrics gradually stabilize, and when $k \geq 5$, its impact on performance becomes marginal.

\subsection{Detailed Results of Different Regularizers.}
\label{app:aggregatio}
\begin{table*}
\caption{Results of different regularizers on the MLLMU 10\% Task, with ours \colorbox{gray!30}{highlighted}. All reported values are percentages (\%).}
\centering
\resizebox{\textwidth}{!}{
\begin{tabular}{c|cc|cc|ccc|ccc}
\toprule
\multicolumn{1}{l}{}          & \multicolumn{2}{|c|}{Forget}                              & \multicolumn{2}{c|}{Generalization}                      & \multicolumn{3}{c|}{Retain}                                                             & \multicolumn{3}{c}{Real}                                                              \\
\midrule
10\% Task        & Acc                          & Gib↑                    & Acc                          & Gib↑                    & Rouge↑                       & BLEU↑                        & Gib↑                    & Rouge↑                       & BLEU↑                       & Gib↑                    \\
\midrule
                        & 25.3                         & 83.7                         & 30.1                         & 78.5                         & 20.6                         & 3.0                          & 81.3                         & 17.5                         & 2.4                          & 87.8                         \\
\multirow{-2}{*}{None}  & ±2.9                         & ±2.1                         & ±2.5                         & ±2.7                         & ±2.9                         & ±0.7                         & ±3.5                         & ±2.1                         & ±0.6                         & ±2.6                         \\
                        & 24.8                         & 88.2                         & 29.5                         & 83.7                         & 23.3                         & 3.6                          & 86.0                         & 18.8                         & 2.7                          & 90.3                         \\
\multirow{-2}{*}{w CoS} & ±3.3                         & ±0.3                         & ±3.4                         & ±0.6                         & ±1.7                         & ±0.7                         & ±0.9                         & ±1.2                         & ±0.6                         & ±0.5                         \\
                        & 30.3                         & 95.0                         & 31.7                         & 85.3                         & 33.3                         & 7.0                          & 93.8                         & 25.0                         & 5.0                          & 93.7                         \\
\multirow{-2}{*}{w JSD} & ±5.1                         & ±1.1                         & ±1.8                         & ±1.5                         & ±1.5                         & ±0.7                         & ±0.4                         & ±1.2                         & ±0.5                         & ±0.4                         \\
                        & \cellcolor[HTML]{E8E8E8}30.4 & \cellcolor[HTML]{E8E8E8}95.4 & \cellcolor[HTML]{E8E8E8}30.1 & \cellcolor[HTML]{E8E8E8}88.2 & \cellcolor[HTML]{E8E8E8}32.4 & \cellcolor[HTML]{E8E8E8}6.9  & \cellcolor[HTML]{E8E8E8}94.6 & \cellcolor[HTML]{E8E8E8}24.6 & \cellcolor[HTML]{E8E8E8}4.7  & \cellcolor[HTML]{E8E8E8}93.7 \\
\multirow{-2}{*}{w KL}  & \cellcolor[HTML]{E8E8E8}±3.6 & \cellcolor[HTML]{E8E8E8}±0.3 & \cellcolor[HTML]{E8E8E8}±1.9 & \cellcolor[HTML]{E8E8E8}±3.3 & \cellcolor[HTML]{E8E8E8}±1.4 & \cellcolor[HTML]{E8E8E8}±0.5 & \cellcolor[HTML]{E8E8E8}±0.4 & \cellcolor[HTML]{E8E8E8}±1.1 & \cellcolor[HTML]{E8E8E8}±0.4 & \cellcolor[HTML]{E8E8E8}±0.2
\\
\bottomrule
\end{tabular}}
\label{tab:regular}
\end{table*}
Table~\ref{tab:regular} provides a quantitative supplement to the results shown in Figure~\ref{fig:aggre}. Consistent with the observations in the main text, the numerical results in Table~\ref{tab:regular} lead to the same conclusions. 

Introducing CoS, compared with the variant without regularization, significantly enhances language coherence, while offering only marginal improvements in unlearning and retention performance. The performance obtained with JSD is almost indistinguishable from that achieved using KL divergence. Nevertheless, in terms of computational cost, JSD requires approximately twice the computation of KL divergence. As a result, employing KL divergence as the regularization term constitutes a more efficient and practical design choice.

\subsection{Detailed Results of Different Visual References.}
\label{app:reference}
\begin{table*}[t]
\caption{Results of different references on the MLLMU 10\% Task, with ours \colorbox{gray!30}{highlighted}. All reported values are percentages (\%).}
\centering
\resizebox{\textwidth}{!}{
\begin{tabular}{c|cc|cc|ccc|ccc}
\toprule
\multicolumn{1}{l}{}          & \multicolumn{2}{|c|}{Forget}                              & \multicolumn{2}{c|}{Generalization}                      & \multicolumn{3}{c|}{Retain}                                                             & \multicolumn{3}{c}{Real}                                                              \\
\midrule
10\% Task        & Acc                          & Gib↑                    & Acc                          & Gib↑                    & Rouge↑                       & BLEU↑                        & Gib↑                    & Rouge↑                       & BLEU↑                       & Gib↑                    \\
\midrule
                          & 29.9                         & 94.9                         & 32.7                         & 91.7                         & 33.8                         & 7.3                          & 95.2                         & 24.9                         & 4.8                          & 94.4                         \\
\multirow{-2}{*}{Forget}  & ±5.0                         & ±0.7                         & ±2.8                         & ±2.0                         & ±2.3                         & ±0.9                         & ±0.1                         & ±1.7                         & ±0.7                         & ±0.4                         \\
                          & 29.9                         & 94.0                         & 29.9                         & 87.0                         & 32.6                         & 6.8                          & 94.3                         & 24.2                         & 4.6                          & 94.1                         \\
\multirow{-2}{*}{Retain}  & ±3.3                         & ±1.2                         & ±2.7                         & ±2.6                         & ±1.4                         & ±0.6                         & ±0.3                         & ±1.1                         & ±0.4                         & ±0.3                         \\
                          & \cellcolor[HTML]{E8E8E8}30.4 & \cellcolor[HTML]{E8E8E8}95.4 & \cellcolor[HTML]{E8E8E8}30.1 & \cellcolor[HTML]{E8E8E8}88.2 & \cellcolor[HTML]{E8E8E8}32.4 & \cellcolor[HTML]{E8E8E8}6.9  & \cellcolor[HTML]{E8E8E8}94.6 & \cellcolor[HTML]{E8E8E8}24.6 & \cellcolor[HTML]{E8E8E8}4.7  & \cellcolor[HTML]{E8E8E8}93.7 \\
\multirow{-2}{*}{People}  & \cellcolor[HTML]{E8E8E8}±3.6 & \cellcolor[HTML]{E8E8E8}±0.3 & \cellcolor[HTML]{E8E8E8}±1.9 & \cellcolor[HTML]{E8E8E8}±3.3 & \cellcolor[HTML]{E8E8E8}±1.4 & \cellcolor[HTML]{E8E8E8}±0.5 & \cellcolor[HTML]{E8E8E8}±0.4 & \cellcolor[HTML]{E8E8E8}±1.1 & \cellcolor[HTML]{E8E8E8}±0.4 & \cellcolor[HTML]{E8E8E8}±0.2 \\
                          & 28.7                         & 91.3                         & 27.6                         & 80.0                         & 30.5                         & 6.3                          & 92.7                         & 24.2                         & 4.5                          & 93.5                         \\
\multirow{-2}{*}{Pets}    & ±4.4                         & ±1.7                         & ±2.8                         & ±4.2                         & ±1.4                         & ±0.7                         & ±0.4                         & ±1.8                         & ±0.7                         & ±1.0                         \\
                          & 31.1                         & 95.5                         & 31.3                         & 87.9                         & 35.0                         & 7.7                          & 94.8                         & 25.5                         & 5.1                          & 94.0                         \\
\multirow{-2}{*}{Scene}   & ±3.3                         & ±0.3                         & ±2.0                         & ±2.7                         & ±1.5                         & ±0.7                         & ±0.3                         & ±1.0                         & ±0.4                         & ±0.0                         \\
                          & 30.1                         & 94.2                         & 30.9                         & 85.9                         & 34.2                         & 7.3                          & 94.4                         & 25.0                         & 4.9                          & 93.9                         \\
\multirow{-2}{*}{Pattern} & ±3.7                         & ±0.6                         & ±2.7                         & ±3.3                         & ±1.5                         & ±0.6                         & ±0.3                         & ±1.2                         & ±0.4                         & ±0.0                        
\\
\bottomrule
\end{tabular}}
\label{tab:ref_10}
\end{table*}

\begin{table*}[!h]
\caption{Results of different references on the MLLMU 15\% Task, with ours \colorbox{gray!30}{highlighted}. All reported values are percentages (\%).}
\centering
\resizebox{\textwidth}{!}{
\begin{tabular}{c|cc|cc|ccc|ccc}
\toprule
\multicolumn{1}{l}{}          & \multicolumn{2}{|c|}{Forget}                              & \multicolumn{2}{c|}{Generalization}                      & \multicolumn{3}{c|}{Retain}                                                             & \multicolumn{3}{c}{Real}                                                              \\
\midrule
15\% Task        & Acc                          & Gib↑                    & Acc                          & Gib↑                    & Rouge↑                       & BLEU↑                        & Gib↑                    & Rouge↑                       & BLEU↑                       & Gib↑                    \\
\midrule
                          & 34.7                         & 93.7                         & 33.1                         & 83.1                         & 49.7                         & 19.0                         & 95.0                         & 31.5                         & 9.5                          & 92.6                         \\
\multirow{-2}{*}{Forget}  & ±0.6                         & ±1.6                         & ±3.0                         & ±2.5                         & ±1.8                         & ±2.4                         & ±0.7                         & ±1.2                         & ±0.6                         & ±1.6                         \\
                          & 30.7                         & 94.3                         & 28.4                         & 84.4                         & 45.8                         & 14.7                         & 95.1                         & 29.6                         & 8.5                          & 93.2                         \\
\multirow{-2}{*}{Retain}  & ±1.2                         & ±1.1                         & ±2.2                         & ±3.4                         & ±1.3                         & ±1.1                         & ±1.1                         & ±0.9                         & ±0.7                         & ±0.8                         \\
                          & \cellcolor[HTML]{E8E8E8}32.0 & \cellcolor[HTML]{E8E8E8}93.2 & \cellcolor[HTML]{E8E8E8}32.4 & \cellcolor[HTML]{E8E8E8}83.1 & \cellcolor[HTML]{E8E8E8}52.7 & \cellcolor[HTML]{E8E8E8}22.0 & \cellcolor[HTML]{E8E8E8}94.6 & \cellcolor[HTML]{E8E8E8}33.2 & \cellcolor[HTML]{E8E8E8}10.3 & \cellcolor[HTML]{E8E8E8}92.0 \\
\multirow{-2}{*}{People}  & \cellcolor[HTML]{E8E8E8}±3.1 & \cellcolor[HTML]{E8E8E8}±0.7 & \cellcolor[HTML]{E8E8E8}±3.4 & \cellcolor[HTML]{E8E8E8}±0.8 & \cellcolor[HTML]{E8E8E8}±1.8 & \cellcolor[HTML]{E8E8E8}±1.7 & \cellcolor[HTML]{E8E8E8}±0.0 & \cellcolor[HTML]{E8E8E8}±0.6 & \cellcolor[HTML]{E8E8E8}±0.2 & \cellcolor[HTML]{E8E8E8}±0.7 \\
                          & 34.5                         & 95.0                         & 29.4                         & 84.4                         & 45.9                         & 14.2                         & 96.0                         & 28.6                         & 8.5                          & 93.8                         \\
\multirow{-2}{*}{Pets}    & ±1.5                         & ±0.2                         & ±1.9                         & ±1.7                         & ±2.6                         & ±1.1                         & ±0.2                         & ±2.4                         & ±1.1                         & ±0.8                         \\
                          & 35.1                         & 94.4                         & 32.3                         & 84.0                         & 50.3                         & 17.9                         & 95.7                         & 31.9                         & 9.8                          & 93.8                         \\
\multirow{-2}{*}{Scene}   & ±0.9                         & ±0.6                         & ±2.5                         & ±1.5                         & ±2.6                         & ±3.4                         & ±0.4                         & ±1.3                         & ±0.6                         & ±1.0                         \\
                          & 31.9                         & 93.8                         & 27.9                         & 84.7                         & 50.4                         & 18.8                         & 95.2                         & 31.9                         & 9.8                          & 93.4                         \\
\multirow{-2}{*}{Pattern} & ±0.5                         & ±0.5                         & ±0.5                         & ±1.5                         & ±2.5                         & ±2.9                         & ±0.5                         & ±0.9                         & ±0.4                         & ±1.5                        
\\
\bottomrule
\end{tabular}}
\label{tab:ref_15}
\end{table*}
Table~\ref{tab:ref_10} and Table~\ref{tab:ref_15} complement Figure~\ref{fig:ref} by reporting results obtained with different types of visual references. From the quantitative results, it can be observed that for relatively easier tasks, such as the 10\% unlearning setting, the performance differences across various visual references are not particularly pronounced. However, for more challenging scenarios, such as the 15\% unlearning task, the impact of reference selection becomes more evident. In this case, conclusions consistent with those in the main text can be drawn: different visual references lead to similar performance on GIB, while noticeable differences emerge in unlearning and retention metrics, while unlearning and retaining performance vary.

Specifically, visual references based on `{pattern}' and `{retain}' images achieve the strongest unlearning performance, which can be attributed to their more stable internal distributions and their substantial distributional gap from the forget set. In contrast, `{forget}' and `{scene}' images perform the worst, as the former directly contains sensitive information targeted for removal, while the latter exhibits high visual diversity, both of which impede accurate estimation of the ideal token distribution. 

Regarding retention performance, the use of `{retain}' and `{pet}' images leads to more severe degradation, possibly because their estimated distributions encode task-relevant or human-like information that is inadvertently suppressed during unlearning. By comparison, `{people}' images provide a more favorable trade-off between effective unlearning and retention.

\end{document}